\documentclass[onecolumn, 10pt]{article}

\usepackage{amssymb}
\usepackage{amsmath}
\usepackage{graphicx}
\usepackage{subfigure}
\usepackage{mathrsfs}
\usepackage{algorithm}
\usepackage{algorithmic}
\usepackage{natbib}

\newtheorem{theorem}{Theorem}
\newtheorem{lemma}{Lemma}
\newtheorem{rem}{Remark}
\author{%
	Cristian Rusu
	\\
	University Politehnica of Bucharest, Romania\\
	\texttt{cristian.rusu@upb.ro} \\
	Lorenzo Rosasco \\
	LCSL, Universit\'a di Genova\\
	Massachusetts Institute of Technology and Istituto Italiano di Tecnologia
}
\title{Constructing fast approximate eigenspaces with application to the fast graph Fourier transforms}
\date{}

\begin{document}

\maketitle

\begin{abstract}
We investigate numerically efficient approximations of eigenspaces associated to symmetric and general matrices. The eigenspaces are factored into a fixed number of fundamental components that can be efficiently manipulated (we consider extended orthogonal Givens or scaling and shear transformations). The number of these components controls the trade-off between approximation accuracy and the computational complexity of projecting on the eigenspaces. We write minimization problems for the single fundamental components and provide closed-form solutions. Then we propose algorithms that iterative update all these components until convergence. We show results on random matrices and an application on the approximation of graph Fourier transforms for directed and undirected graphs.
\end{abstract}

\section{Introduction}

Matrix decomposition techniques \citep{814658}, and specifically eigenvalue decompositions \citep{DecompositionsEig}, are widely used in numerical linear algebra, scientific computing, machine learning, quantum computing and other scientific fields.

In general, given no assumptions on the structure of an eigenspace, the eigenvector matrix of a given linear operator of size $n \times n$ exhibits no advantageous numerical properties and therefore they require $O(n^2)$ operations when performing matrix-vector multiplications. In this paper, we want to perform an approximate eigenvalue decomposition so that we accurately represent the original eigenspace with a new one that also exhibits favorable numerical complexity, for example, it requires only $O(n \log n)$ operations when performing matrix-vector multiplication with a generic vector. In such cases, a trade-off between the accuracy of the approximation and its numerical complexity exists. Several previous works, such as \citep{Treelets} and \citep{Kondor2014MMF}, have already introduced these ideas in the machine learning community with considerable success. Such approximations are particularly useful in situations where, once computed, the eigenspace is repeatedly used in matrix-vector calculations in downstream applications.

Eigendecomposition algorithms developed in the matrix computations literature are different for symmetric and unsymmetric matrices. In the symmetric case, the eigenspace is always full (non-defective), real-valued and furthermore, orthonormal \citep{Golub1996}[Chapter 8]. We approximate these eigenspaces by using extended Givens transformations (which are themselves orthonormal and include as a particular case the well-known Givens, sometimes also called Jacobi, rotations \citep{GivensRotations}). In this case, given the spectrum or an estimation of it, we can provide a locally optimal iterative algorithm similar to Jacobi diagonalization for symmetric matrices \citep{JacobiProcess}. The general, unsymmetric, case \citep{Golub1996}[Chapter 7] is much more challenging as the given matrix might not even be diagonalizable and furthermore, even when it is, the factorization has to be done over the complex-valued field in general. As the eigenvector matrix is generally unstructured, in this case, we rely on a given number of scaling and shear transformations to approximate it \citep{rusu2018learning}. We formulate optimization problems for each of these basic components and show how to locally, optimally solve them with an iterative algorithm and closed-form solutions.

For both proposed algorithms, we show experimental results on the approximation of random unstructured symmetric matrices and then show an application to the construction of fast graph Fourier transforms on synthetic and real-world directed and undirected graphs.

\section{Prior approaches}

The literature has always distinguished between eigendecompositions of symmetric and unsymmetric matrices and we will do the same.

In the symmetric case, the diagonalization is done with orthonormal matrices, which are well understood in terms of their decomposition with Givens rotations or Householder reflectors (the QR algorithm, see Chapters 5.1 and 5.2 of \citep{Golub1996}). The starting point for some approaches in the literature is the Jacobi diagonalization process for symmetric matrices \citep{JacobiProcess} which is an iterative procedure that uses Givens rotations to bring the symmetric matrix to a strongly diagonally dominant one. A truncated Jacobi procedure is used in \citep{lemagoarou:hal-01416110} to compute fast graph Fourier transforms for undirected graphs (as undirected implies symmetry in the adjacency and Laplacian matrices). Other methods deal directly with the orthonormal eigenspace. For example, Treelets \citep{Treelets} and multiresolution \citep{Kondor2014MMF} structures use Givens rotations in a structured way to decompose the orthonormal components into hierarchies or multiple different scales, respectively. Another approach is to exploit manifold optimization techniques to find approximate factorizations of orthonormal matrices with few Givens rotations either by greedy coordinate descent \citep{Shalit} or by $\ell_1$--style optimization \citep{Frerix2019ApproximatingOM}. Recently, an approach that combines rotations and reflections was proposed with an application to fast principal component analysis (PCA) projections \citep{FastPCA}. While this latter work needs to precompute the orthonormal eigenspace, in this paper we show how to perform the same factorization given the dataset.

In the unsymmetric case, we rely on sparse structured components. For example, the incomplete LU \citep{ILU}, the randomized LU \citep{RLU} factorizations, the additive low-rank plus multiresolution decomposition \citep{mudrakarta2019asymmetric}, and approximate Gaussian elimination \citep{AGE} all rely on structured sparse matrices to construct efficient approximations of a given unstructured matrix.

In this paper, we use structured matrices to construct numerically efficient approximations of eigenspaces. We describe the fundamental building blocks of our factorizations and provide exact optimization problems with closed-form solutions to optimally, locally update these blocks efficiently.

\section{Problem setup and formulation}

\subsection{The symmetric case}


Given a symmetric matrix $\mathbf{S} \in \mathbb{R}^{n \times n}$ the main result that we use is its eigenvalue factorization as
\begin{equation}
\mathbf{S} = \mathbf{U}\text{diag}(\mathbf{s})\mathbf{U}^T,\ \mathbf{UU}^T = \mathbf{U}^T \mathbf{U} = \mathbf{I}, \ \mathbf{s} \in \mathbb{R}^n,
\label{eq:symmetric}
\end{equation}
where we assume w.l.o.g. that the entries of $\mathbf{s}$, which are the real-valued eigenvalues of $\mathbf{S}$, are in descending algebraic order and $\mathbf{U}$ is the real-valued orthonormal eigenspace. Based on \eqref{eq:symmetric}, we consider the problem
\begin{equation}
\underset{\mathbf{\bar{s}},\ \mathbf{\bar{U}}}{\text{minimize}} \ \| \mathbf{S} - \mathbf{\bar{U}} \text{diag}(\mathbf{\bar{s}}) \mathbf{\bar{U}}^T \|_F^2 \text{ subject to } \mathbf{\bar{U}}\! \in \! \mathcal{G}_g,
\label{eq:symmetricproblem}
\end{equation}
where $\mathcal{G}_g$ a set of orthonormal matrices such that matrix-vector multiplication with any matrix from this set is $O(g)$, instead of the classic $O(n^2)$. Let us now consider a particular set $\mathcal{G}_g$. Based on all the $2 \times 2$ orthonormal matrices
\begin{equation}
\mathbf{\tilde{G}}  \in \left\{ \begin{bmatrix}
c & s \\
-s & c
\end{bmatrix},\ \begin{bmatrix}
c & s \\
s & -c
\end{bmatrix} \right\},\ c^2+s^2 = 1,
\label{eq:localstructure}
\end{equation}
we have the extended orthonormal Givens transformations \citep{FastSparsifyingTransforms, FastPCA}, which for simplicity we call a G-transform:
\begin{equation}
\mathbf{G}_{ij} = \begin{bmatrix} 
\mathbf{I}_{i-1} &  &  &  & \\
& * &  & * & \\
& & \mathbf{I}_{j-i-1} & & \\
& * & & * & \\
& & & & \mathbf{I}_{n-j} \\
\end{bmatrix} \in \mathbb{R}^{n \times n},
\label{eq:Gtransform}
\end{equation}
where the non-zero entries located at rows/columns $i$ and $j$, denoted as ``*", are the two possible options in \eqref{eq:localstructure}. The matrices in \eqref{eq:localstructure} are basic building blocks of the orthonormal group because every orthonormal matrix $\mathbf{U} \in \mathbb{R}^{n \times n}$ can be diagonalized (and the diagonal entries are $\{ \pm 1 \}$) using $\frac{n(n-1)}{2}$ Givens rotations (the matrix \eqref{eq:Gtransform} with the first, unsymmetric, component in \eqref{eq:localstructure}) by the QR decomposition of $\mathbf{U}$, see Chapter~5.2.5 of \citep{Golub1996}. Then, in this paper, any $\mathbf{\bar{U}} \in \mathcal{G}_g$ has the following structure
\begin{equation}
\mathbf{\bar{U}} = \prod_{k=1}^g \mathbf{G}_{i_k j_k} = \mathbf{G}_{i_g j_g} \dots \mathbf{G}_{i_2 j_2} \mathbf{G}_{i_1 j_1},
\label{eq:Ubar}
\end{equation}
where all matrices $\mathbf{G}_{i_k j_k}$ are G-transforms \eqref{eq:Gtransform}. The number $g \ll n^2$ is given and fixed. With this structure, matrix-vector multiplication $\mathbf{\bar{U}x}$ takes $6g$ operations while storing $\mathbf{\bar{U}}$ takes approximately $2g \log_2 n + gC$ bits, where $C$ is the number of bits required for a double precision floating-point representation. Similar structures to \eqref{eq:Ubar} have been previously proposed by \citep{Treelets}, \citep{Kondor2014MMF}, and \citep{Frerix2019ApproximatingOM}, but they all consider only Givens rotations, and no reflectors.

\subsection{The unsymmetric case}

Given a general diagonalizable $\mathbf{C} \in \mathbb{R}^{n \times n}$ the main result that we use is its eigenvalue factorization as
\begin{equation}
\mathbf{C} = \mathbf{T} \text{diag}(\mathbf{c}) \mathbf{T}^{-1}, \ \mathbf{c} \in \mathbb{C}^n,
\label{eq:unsymmetric}
\end{equation}
where $\mathbf{c}$ contains the complex-valued eigenvalues and $\mathbf{T}$ has the complex-valued eigenvectors. Based on \eqref{eq:unsymmetric}, we consider the problem
\begin{equation}
\underset{\mathbf{\bar{c}},\ \mathbf{\bar{T}}}{\text{minimize}} \ \| \mathbf{C} \! -\! \mathbf{\bar{T}} \text{diag}(\mathbf{\bar{c}}) \mathbf{\bar{T}}^{-1}\! \|_F^2 \text{ subject to } \mathbf{\bar{T}}\! \in \! \mathcal{T}_m,
\label{eq:unsymmetricproblem}
\end{equation}
where $\mathcal{T}_m$ a set of general matrices such that matrix-vector multiplication with any matrix from this set or its inverse is $O(m)$, instead of $O(n^2)$. Let us now consider a particular set $\mathcal{T}_m$. Based on $2 \times 2$ scaling and shear transformations
\begin{equation}
\mathbf{\tilde{T}} \in \left\{\begin{bmatrix}
a & 0 \\
0 & 1
\end{bmatrix},
\begin{bmatrix}
1 & a \\
0 & 1
\end{bmatrix}, \begin{bmatrix}
1 & 0 \\
a & 1
\end{bmatrix}  \right\},\ a \in \mathbb{R},
\label{eq:localstructure2}
\end{equation}
and, similarly to \eqref{eq:Gtransform}, we define the T-transform:
\begin{equation}
\mathbf{T}_{ij} = \begin{bmatrix} 
\mathbf{I}_{i-1} &  &  &  & \\
& * &  & * & \\
& & \mathbf{I}_{j-i-1} & & \\
& * & & * & \\
& & & & \mathbf{I}_{n-j} \\
\end{bmatrix} \in \mathbb{R}^{n \times n},
\label{eq:Ttransform}
\end{equation}
where the non-zero entries are the three possible options in \eqref{eq:localstructure2}. For the shear transformations we necessarily have $j>i$ while for the the scaling transformations we abuse notation and impose $i=j$, i.e., $\mathbf{T}_{ii} = \mathbf{T}_i$ in \eqref{eq:Ttransform} is the identity matrix except for the $i^\text{th}$ diagonal element that is $a$. The matrices in \eqref{eq:Ttransform} are building blocks of every diagonalizable $\mathbf{C} \in \mathbb{R}^{n \times n}$ because, by Gaussian elimination (see Chapter~3.2.1 of \citep{Golub1996}), $n^2-n$  shear transformations \eqref{eq:Ttransform} diagonalize $\mathbf{C}$ and then $n$ scaling transformations \eqref{eq:Ttransform} exactly represent the resulting diagonal. We choose these three specific matrix as the optimization in \eqref{eq:unsymmetricproblem} takes place over the variable $\mathbf{\bar{T}}$ and its inverse and these matrices have trivial inverses. Then, in this paper, any $\mathbf{\bar{T}} \in \mathcal{T}_m$ has the following structure
\begin{equation}
\mathbf{\bar{T}} = \prod_{k=1}^m \mathbf{T}_{i_k j_k} = \mathbf{T}_{i_m j_m} \dots \mathbf{T}_{i_2 j_2} \mathbf{T}_{i_1 j_1},
\label{eq:Tbar}
\end{equation}
where all matrices $\mathbf{T}_{i_k j_k}$ are T-transforms \eqref{eq:Ttransform}. Their number $m \ll n^2$ is given and fixed. We assume that the factorization contains $m_1$ scalings and $m_2$ shears ($m_1 + m_2 = m$). With this structure, matrix-vector multiplication $\mathbf{\bar{T}x}$ takes $m_1 + 2m_2$ operations while storing $\mathbf{\bar{T}}$ takes approximately $m C + (m_1 + 2m_2) \log_2 n$ bits, where $C$ is the number of bits required for a double precision floating-point representation.

There are two important differences when compared to G-transforms. Neither the scaling nor the shears are orthogonal but they are more efficient: two computations and one degree of freedom per transform (as compared to the G-transform where we have 6 operations and one degree of freedom). Therefore, for the same computational cost, we expect T-transforms to provide more accurate approximations. The two types of transforms are connected since any $2 \times  2$ orthonormal transformation can be written as a product of three shears and scalings by the lifting scheme \citep{Lifting2}.

\section{Proposed factorizations and algorithms}

In this section we propose approximate solutions to the optimization problems \eqref{eq:symmetric} and \eqref{eq:unsymmetric}. Therefore, we distinguish between the symmetric and unsymmetric cases. Furthermore, we analyze separately the initialization and iterative procedures that improve the approximation for each of the two problems. Both \eqref{eq:symmetricproblem} and \eqref{eq:unsymmetric} echo the fast circulant matrix-vector multiplication which is possible because every circulant matrix of size $n \times n$ has a factorization as $\mathbf{F}^H \mathbf{\Sigma F}$, Chapter~4.8 of \citep{Golub1996}, where $\mathbf{F}$ is the Fourier matrix and $\mathbf{\Sigma} = \text{diag}(\sigma)$, $\mathbf{\sigma} \in \mathbb{C}^n$. The idea is to replace the Fourier matrix with a new learned matrix ($\mathbf{\bar{U}}$ as \eqref{eq:Ubar} or $\mathbf{\bar{T}}$ as \eqref{eq:Tbar}), with similar computational properties to the Fourier \citep{FFT}.

\subsection{Approximation of symmetric matrices}

Based on the eigenvalue decomposition, the idea is to approximate $\mathbf{U}$ in \eqref{eq:symmetric} with a fast approximation $\mathbf{\bar{U}}$ \eqref{eq:Ubar}. The approximation of $\mathbf{S}$ would therefore be $\mathbf{\bar{S}} = \mathbf{\bar{U} \text{diag}(\mathbf{\bar{s}}) \bar{U}}^T$. Multiplications with $\mathbf{S}$ can be viewed as a sequence of fast multiplications by $\mathbf{\bar{U}}^T$, $\text{diag}(\mathbf{s})$ and finally $\mathbf{\bar{U}}$. Matrix-vector multiplication with a diagonal matrix is fast, $n$ operations, so therefore our goal is to construct $\mathbf{\bar{U}}$ such that it also has advantageous numerical properties (for example, computations take less than $2n^2$ operations). Therefore, based on \eqref{eq:Ubar}, we propose an approximation as
\begin{equation}
\mathbf{\bar{S}} = \left( \prod_{k=1}^g \mathbf{G}_{i_k j_k} \right) \text{diag}(\mathbf{\bar{s}}) \left( \prod_{k=g}^{1} \mathbf{G}_{i_k j_k}^T \right), \ \mathbf{\bar{s}} \in \mathbb{R}^n,
\label{eq:approximations4}
\end{equation}
where $\mathbf{\bar{s}}$ are the estimated eigenvalues of $\mathbf{S}$. These can be the actual eigenvalues of $\mathbf{S}$ if they are known, or else they can be randomly initialized or set to the diagonal elements of $\mathbf{S}$ -- in the latter case, we should ensure entries of $\mathbf{\bar{s}}$ are distinct for reasons that will be clear later in this section. To find the best approximation $\mathbf{\bar{S}}$ of $\mathbf{S}$ we can compute the best approximation to the spectrum by the following lemma.
\begin{lemma}
    Let $\mathbf{S}$ and orthogonal $\mathbf{\bar{U}}$ be fixed, then $\mathbf{\bar{s}}^\star$ the $\arg \min$ of the expression $\|  \mathbf{S} - \mathbf{\bar{U}} \text{diag}(\mathbf{\bar{s}}) \mathbf{\bar{U}}^T \|_F^2$ is given by
    \begin{equation}
    \mathbf{\bar{s}}^\star = \text{diag}( \mathbf{\bar{U}}^T \mathbf{S} \mathbf{\bar{U}}  ).
    \end{equation}
\end{lemma}
The complexity of computing $\mathbf{\bar{s}}^\star$ is $O(gn)$.

Let us now move to approximate the orthogonal eigenspace of $\mathbf{S}$. Given an approximation as \eqref{eq:approximations4} where all G-transforms $t = k+1,\dots, g$ were initialized, we now study the problem of initializing $\mathbf{G}_{i_k j_k}$ such that we minimize
\begin{equation}
\| \mathbf{S} - \mathbf{\bar{S}} \|_F^2 = \| \mathbf{S}^{(k)}  - \mathbf{G}_{i_k j_k}\text{diag}(\mathbf{\bar{s}}) \mathbf{G}_{i_k j_k}^T \|_F^2,
\label{eq:initializationsymmetricWithGivens}
\end{equation}
where we have defined the symmetric matrix
\begin{equation}
\mathbf{S}^{(k)} = \left( \prod_{t=g}^{k+1} \mathbf{G}_{i_t j_t} \right) \mathbf{S} \left( \prod_{t=k+1}^{g} \mathbf{G}_{i_t j_t}^T \right).
\label{eq:sk}
\end{equation}

\begin{theorem}\textbf{(Optimal initialization of each G-transform)} Let $\mathbf{S}$, $\mathbf{\bar{s}}$ be fixed and all components $\mathbf{G}_{i_t j_t} \neq \mathbf{I}_n$ for $t = k+1,\dots,g$ while $\mathbf{G}_{i_t j_t} = \mathbf{I}_n$ for $t=1,\dots,k-1$, then the optimal $k^\text{th}$ component $\mathbf{G}_{i_k^\star j_k^\star} = \arg \min \| \mathbf{S}^{(k)}  - \mathbf{G}_{i_k j_k}\text{diag}(\mathbf{\bar{s}}) \mathbf{G}_{i_k j_k}^T  \|_F^2$ has its non-trivial values given by $\mathbf{\tilde{G}}_{k} = \mathbf{V}_k^T \text{ with } \mathbf{S}^{(k)}_{ \{i_k^\star, j_k^\star \} } = \mathbf{V}_k \mathbf{D}_k \mathbf{V}_k^T$ for the optimal coordinates
    \begin{equation}
    (i_k^\star, j_k^\star) = \underset{(i,j),\ j > i}{\arg \max} \ \mathscr{A}_{i j} \text{ with } \mathscr{A}_{i j} = \gamma_{ i j } (\bar{s}_{j j} - \bar{s}_{i i}),
    \label{eq:scriptA}
    \end{equation}
    where we have denoted the quantity
    \begin{equation}
    \gamma_{ i j } \! \! = \! \! \frac{1}{2} \! \left( \! \! S^{(k)}_{i i } \! \! - \! S^{(k)}_{ j j } \!  + \! \! \sqrt{ \! \left(S^{(k)}_{i i } \! \! - \! \! S^{(k)}_{ j j } \right)^2\! \! \! \! + \! 4\left(\! S^{(k)}_{i j}\! \right)^2} \right),
    \label{eq:thegamma}
    \end{equation}
    and the $2 \times 2$ symmetric $\mathbf{S}^{(k)}_{\{i_k, j_k\} } = \begin{bmatrix}
    S^{(k)}_{i_k i_k} & S^{(k)}_{i_k j_k} \\
    S^{(k)}_{j_k i_k} & S^{(k)}_{j_k j_k}
    \end{bmatrix}$.
\end{theorem}

Theorem 1 provides an efficient way to find the optimal G-transform that minimizes \eqref{eq:initializationsymmetricWithGivens}: both the indices and the transform values. Starting from $k = g$ we can continue down to $k=1$ and initialize in this fashion all $g$ G-transforms in \eqref{eq:Ubar}. Also, notice that the unified approach we propose (allowing for both the rotation and the reflection) simplifies the results, i.e., we are not looking to optimize an angle of rotation but we get the optimal local solution by an eigenvalue decomposition (the solution to a two-sided $2 \times 2$ Procrustes problem). The computational cost of \eqref{eq:scriptA} is dominated by the sweep of the indices ($O(n^2)$ operations).
\begin{rem}\textbf{(Connection to the Jacobi method)} The Jacobi method, see Chapter 8.4 of \citep{Golub1996}, used to diagonalize symmetric matrices, only uses Givens rotations and selects indices $(i_k, j_k)$ that correspond to the off-diagonal element of $\mathbf{S}^{(k)}$ with the highest magnitude, i.e., we have \eqref{eq:scriptA} with $\mathscr{A} = | S_{ij}^{(k)} |$. The Jacobi algorithm is not concerned with the number of Givens rotations used to diagonalize and indeed, in general, more than $n^2$ rotations are used (see Chapter 8.4.3 of \citep{Golub1996} for a detailed discussion on how the number of rotations relates to the converge of the method). Furthermore, the Jacobi method uses only Givens rotations (while we have a richer structure for $\mathbf{G}_{i_k j_k}$ given in \eqref{eq:localstructure}) and it does not explicitly have an objective function as \eqref{eq:symmetricproblem}, i.e., there is no reference matrix to be reconstructed in the sense of minimizing a Frobenius norm (the Jacobi method minimizes the squared sum of the off-diagonal entries of the approximation). Also, the Jacobi method does not need an estimate of the eigenvalues $\mathbf{\bar{s}}$. If we now ignore the eigenvalue information, we can consider $\mathscr{A}_{ij} = \gamma_{ij}$ in \eqref{eq:scriptA}. When $S_{ij}^{(k)} \gg |S^{(k)}_{i i } - S^{(k)}_{ j j }|$ we have that $\mathscr{A}_{ij} \approx | S_{ij}^{(k)} |$, just as in the Jacobi method, while when $S_{ij}^{(k)} \ll |S^{(k)}_{i i } - S^{(k)}_{ j j }|$ we have $\mathscr{A}_{ij} \approx S_{ii}^{(k)} - S_{jj}^{(k)}$. Therefore, the calculated score approximates the Jacobi approach when off-diagonal elements are large but the selection criterium for the indices is different as the iterative process makes progress and the working matrix becomes diagonally dominant.
    Finally, we note that $\mathcal{A}_{ij} = 0$ whenever $\bar{s}_{ii} = \bar{s}_{jj}, \ i \neq j$ which agrees with theoretical convergence results on the Jacobi method that hold when assuming distinct eigenvalues \citep{10.2307/2098731}.
\end{rem}


The proposed approach is also significantly different from other previous approaches. As opposed to the approach in \citep{Kondor2014MMF}, by maximizing \eqref{eq:scriptA} we find the indices of the optimal $2 \times 2$ transform without actually explicitly having to compute it. This saves up computational time (the $\mathscr{A}$ are easy to compute: only 10 operations) and also leads to better approximation (as we also consider the reflector simultaneously with the Givens rotation in \eqref{eq:localstructure}). The approach in \citep{Frerix2019ApproximatingOM} uses again only Givens rotations to perform coordinate descent on the orthonormal manifold using a particular basis for the tangent space such that the exponential map is a Givens rotation. Noting that $\begin{bmatrix}
c & s \\
-s & c
\end{bmatrix} $ has the same structure as $ \begin{bmatrix}
c & s \\
s & -c
\end{bmatrix}  \begin{bmatrix} 0 & 1 \\ 1 & 0 \end{bmatrix}$, i.e., the reflection can be seen as a coordinate swap followed by a rotation, we can interpret our approach as a simultaneous dual tangent space descent. Unfortunately, efforts to integrate this view with manifold optimization, in general, seem difficult at this stage as many difficulties arise: for example, the logarithmic map of the reflector is complex-valued and therefore it is not clear how to choose a basis for the tangent space corresponding to the reflector. As rotations and reflections are disconnected components the unified approach used in this paper may work only for the objective function and constraints we consider (due to the existence of the Procrustes solutions based on eigendecompositions).

Given an approximation as \eqref{eq:approximations4} where all G-transforms were initialized, we now study the problem of improving each individual $\mathbf{G}_{i_k j_k}$ iteratively. We want to optimize each G-transform $\mathbf{G}_{i_k j_k}$ sequentially such that we minimize
\begin{equation}
\| \mathbf{S} - \mathbf{\bar{S}} \|_F^2 = \| \mathbf{A}^{(k)}  - \mathbf{G}_{i_k j_k} \mathbf{B}^{(k)}  \mathbf{G}_{i_k j_k}^T \|_F^2,
\label{eq:symmetricWithGivens}
\end{equation}
where, due to the invariance of the Frobenius norm to multiplications by $\mathbf{G}_{ij}$ and its transpose, we have defined the symmetric matrices
\begin{equation}
\mathbf{A}^{(k)} = \left( \prod_{t={k-1}}^1 \mathbf{G}_{i_t j_t}^T \right) \mathbf{S} \left( \prod_{t=1}^{k-1} \mathbf{G}_{i_t j_t} \right),
\label{eq:ak}
\end{equation}
\begin{equation}
\mathbf{B}^{(k)} = \left( \prod_{t=k+1}^g \mathbf{G}_{i_t j_t} \right) \text{diag}(\mathbf{\bar{s}}) \left( \prod_{t=g}^{k+1} \mathbf{G}_{i_t j_t}^T \right).
\label{eq:bk}
\end{equation}
Notice that Theorem 1 covered only the case when $\mathbf{B}^{(k)}$ is a diagonal matrix. Next we state a general result that holds for the minimization of \eqref{eq:symmetricWithGivens} and any $\mathbf{A}^{(k)}$ and $\mathbf{B}^{(k)}$.

\begin{theorem}\textbf{(Optimal update of each G-transform)} Let $\mathbf{A}^{(k)}$ and $\mathbf{B}^{(k)}$ be any symmetric $n \times n$ matrices, then the minimizer of the quantity in \eqref{eq:symmetricWithGivens} has its non-trivial values given by
    \begin{equation}
    \mathbf{x}^{(f_k)} = \begin{bmatrix} c_{i_k^\star j_k^\star}^\star \\ s_{i_k^\star j_k^\star}^\star \end{bmatrix} = -(\mathbf{R}^{(f_k)} + \lambda^{(f_k)} \mathbf{I}_2)^{-1} \mathbf{g}^{(f_k)},
    \label{eq:getthex}
    \end{equation}
    where $\lambda^{(f_k)} = \min\ \{ \lambda_i \},\text{ where } \mathbf{M}^{(f_k)} \mathbf{v}_i = \lambda_i \mathbf{N}^{(f_k)} \mathbf{v}_i,$ for the optimal coordinates
    \begin{equation}
    (f_k^\star, i_k^\star, j_k^\star) = \underset{f_k \in \{1,2\},\ j_k > i_k}{\arg \min} \mathscr{B}_{i_k j_k}^{(f_k)},
    \end{equation}
    where $\mathscr{B}_{i_k j_k}^{(f_k)} =  (\mathbf{x}^{(f_k)})^T \mathbf{R}^{(f_k)} \mathbf{x}^{(f_k)} + 2(\mathbf{x}^{(f_k)})^T \mathbf{g}^{(f_k)} + \| \mathbf{w} \|_2^2$. The new index $f_k$ runs through the two options (rotation and reflector) in \eqref{eq:localstructure}. The matrices $\mathbf{R}^{(f_k)}$ of size $2 \times 2$, the vectors $\mathbf{g}^{(f_k)}$ of length $2$ and $\mathbf{w}$ of length $n^2$, and the matrices $\mathbf{M}^{(f_k)}$ and $\mathbf{N}^{(f_k)}$ all of size $4 \times 4$ depend only on the entries in $\mathbf{A}^{(k)}$, $\mathbf{B}^{(k)}$ and are given explicitly in the supplementary materials.
\end{theorem}

Unlike with the initialization procedure, Theorem 2 shows that considering both the rotation and the reflector simultaneously does not lead to a unified optimization problem. Indeed, the index $f_k$ runs through both transformations from \eqref{eq:localstructure}. Still, solving the second problem, for $f_k = 2$, brings an extra computational load that is negligible as it shares most calculations with the first problem, for $f_k =1$.

The iterative process is computationally expensive as it covers all $O(n^2)$ unique pairs of indices $(i_k,j_k)$ while the calculation of \eqref{eq:getthex} is itself non-trivial and requires $O(n^3)$ operations (substantially more expensive than the initialization \eqref{eq:thegamma}). If the running time is an important constraint, we can run the iterative process just as a ``polishing step'': keep the indices of the G-transforms fixed all the time and update only the values of the transformations $\mathbf{\tilde{G}}^\star_{k}$.



\subsection{Approximation of unsymmetric matrices}

Similarly to the symmetric case, based now on the eigenvalue decomposition \eqref{eq:unsymmetric}, the idea is to approximate $\mathbf{T}$ with a numerically efficient approximation $\mathbf{\bar{T}}$. The approximation of $\mathbf{C}$ would therefore be $\mathbf{\bar{C}} = \mathbf{\bar{T} \text{diag}(\mathbf{\bar{c}}) \bar{T}}^{-1}$. Multiplications with $\mathbf{\bar{C}}$ can be viewed as a sequence of fast multiplications by $\mathbf{\bar{T}}^{-1}$, $\text{diag}(\mathbf{\bar{c}})$ and finally $\mathbf{\bar{T}}$. Again, matrix-vector multiplication with a diagonal is fast and therefore the computational burden depends on the numerical properties of $\mathbf{\bar{T}}$ and its inverse. By using scaling and shear transformations \eqref{eq:localstructure2} in the direct transformation $\mathbf{\bar{T}}$ the numerical properties transfer also to its inverse $\mathbf{\bar{T}}^{-1}$ since inverses of scalings and shears are themselves scalings and shears, respectively. Therefore, based on \eqref{eq:Tbar}, we propose an approximation as
\begin{equation}
\mathbf{\bar{C}} \! = \! \left( \prod_{k=1}^m \mathbf{T}_{i_k j_k} \right) \! \text{diag}(\mathbf{\bar{c}}) \left( \prod_{k=m}^{1} \mathbf{T}_{i_k j_k}^{-1} \right), \ \mathbf{\bar{c}} \in \mathbb{R}^n,
\label{eq:approximationGeneral}
\end{equation}
where $\mathbf{\bar{c}}$ are the estimated eigenvalues of $\mathbf{C}$, which we constraint to be real-valued. Just like in the symmetric case, there are several ways to set $\mathbf{\bar{c}}$: randomly, the diagonal values of $\mathbf{C}$ or the true eigenvalues, if they are known. To find the best approximation $\mathbf{\bar{C}}$ of $\mathbf{C}$ we can compute the best approximation to the spectrum by the following lemma.
\begin{lemma}
    Let $\mathbf{C}$ and $\mathbf{\bar{T}}$ be fixed, then $\mathbf{\bar{c}}^\star$ the $\arg \min$ of the expression $\|  \mathbf{C} - \mathbf{\bar{T}} \text{diag}(\mathbf{\bar{c}}) \mathbf{\bar{T}}^{-1} \|_F^2$ is given by
    \begin{equation}
    \mathbf{\bar{c}}^\star = (\mathbf{\bar{T}}^{-T} * \mathbf{\bar{T}} )^{-1} \text{vec}(\mathbf{C}),
    \end{equation}
    where $*$ is the Khatri-Rao product.
\end{lemma}
The computational complexity of getting $\mathbf{\bar{c}}^\star$ is $O(n^4)$, we solve a least squares problem of size $n^2 \times n$ where the columns of $\mathbf{\bar{T}}^{-T} * \mathbf{\bar{T}} $ are Kronecker products. Alternatively, we can approximately solve the problem by some iterative method which exploits the Kronecker product structure to efficiently perform matrix-vector multiplications.

Let us now move to approximate the eigenspace of $\mathbf{T}$. Given an approximation as \eqref{eq:approximationGeneral} where all T-transforms $t = 1,\dots,k-1$ were initialized, we now study the problem of initializing $\mathbf{T}_{i_k j_k}$ such that we minimize
\begin{equation}
\| \mathbf{C} - \mathbf{\bar{C}} \|_F^2 = \| \mathbf{C}  - \mathbf{T}_{i_k j_k} \mathbf{B}^{(k)} \mathbf{T}_{i_k j_k}^{-1} \|_F^2,
\label{eq:initializationGeneral}
\end{equation}
where we have defined the symmetric matrix
\begin{equation}
\mathbf{B}^{(k)} = \left( \prod_{t=1}^{k-1} \mathbf{T}_{i_t j_t} \right) \text{diag}(\mathbf{\bar{c}}) \left( \prod_{t=k-1}^{1} \mathbf{T}_{i_t j_t}^{-1} \right).
\label{eq:theNewbk}
\end{equation}



\begin{theorem}\textbf{(Optimal initialization of each T-transform)} Let $\mathbf{C}$ and $\mathbf{\bar{c}}$ be fixed, let all components $\mathbf{T}_{i_t j_t} \neq \mathbf{I}_n$ for $t=1,\dots,k-1$ while $\mathbf{T}_{i_t j_t} = \mathbf{I}_n$ for $t = k+1,\dots,m$, then the optimal $k^\text{th}$ component $\mathbf{T}_{i_k^\star j_k^\star}^\star$ that minimizes the quantity in \eqref{eq:initializationGeneral} is given by
    \begin{equation}
    (f_k^\star, i_k^\star, j_k^\star, a_k^\star) = \underset{f_k \in \{1,2,3 \},\ j_k > i_k}{\arg \min} \mathscr{C}_{i_k j_k}^{(f_k)}(a_k),
    \label{eq:optC}
    \end{equation}
    where the quantities $\mathscr{C}_{i_k j_k}^{(f_k)}(a_k)$, for an index $f_k$ that runs through all three options in \eqref{eq:localstructure2}, are rational functions in $a_k$ and are given explicitly in the supplementary materials.
\end{theorem}

Given an approximation as \eqref{eq:approximationGeneral} where all T-transforms were initialized, we now study the problem of improving each individual $\mathbf{T}_{i_k j_k}$ iteratively. Therefore, we want to optimize each T-transform $\mathbf{T}_{i_k j_k}$ sequentially such that we minimize
\begin{equation}
\| \mathbf{C} \! - \mathbf{\bar{C}} \|_F^2 \! =\! \| \mathbf{C}\!  - \mathbf{A}^{(k)} \mathbf{T}_{i_k j_k} \mathbf{B}^{(k)}  \mathbf{T}_{i_k j_k}^{-1} (\mathbf{A}^{(k)})^{-1}\! \|_F^2,
\label{eq:unsymmetricWithT}
\end{equation}
where we have defined the matrix
\begin{equation}
\mathbf{A}^{(k)} = \prod_{t={k+1}}^m \mathbf{T}_{i_t j_t}.
\end{equation}


\begin{theorem}\textbf{(Optimal update of each T-transform)} Let $\mathbf{A}^{(k)}$ and $\mathbf{B}^{(k)}$ be any $n \times n$ matrices, then the optimal $k^\text{th}$ component $\mathbf{T}_{i_k^\star j_k^\star}^\star$ that minimizes the quantity in \eqref{eq:unsymmetricWithT} is given by
    \begin{equation}
    (f_k^\star, i_k^\star, j_k^\star, a_k^\star) = \underset{f_k \in \{1,2,3 \},\ j_k > i_k}{\arg \min} \mathscr{D}_{i_k j_k}^{(f_k)}(a_k),
    \label{eq:optD}
    \end{equation}
    where the quantities $\mathscr{D}_{i_k j_k}^{(f_k)}(a_k)$, for an index $f_k$ that runs through all three options in \eqref{eq:localstructure2}, are rational functions in $a_k$ and are given explicitly in the supplementary materials.
\end{theorem}

Although the initialization and iterative steps look very similar, \eqref{eq:optC} and \eqref{eq:optD}, they are significantly different from a computational perspective. While a particular $\mathscr{C}_{i_k j_k}^{(f_k)} (a_k)$ is computed in constant time $O(1)$, for each of the $\mathscr{D}^{(f_k)} (a_k)$s we have quadratic complexity $O(n^2)$.

Similarly to the symmetric case, we now have a locally optimal way of choosing and updating our T-transforms. Also, the update step is again significantly more computationally expensive than the initialization and more are more expensive than the results for the symmetric case. The basic difficulty stems from the fact that we are dealing now with building blocks that are not orthogonal and therefore are not invariant in the Frobenius norm. The exact initialization of T-transforms takes $O(n^3)$ while the update takes $O(n^4)$ operations. Due to this high computational cost, simplification can be brought to the algorithm: for example, in the update steps we no longer search over every index pair $(i_k,j_k)$ but we keep these indices fixed and just calculate the locally optimal coefficient of the transformation $a_k$. We call this step a polishing step and it reduces the computational complexity of updating the T-transforms to $O(n^3)$.

We now make two remarks regarding the proposed decomposition for general matrices.

\begin{algorithm}[!t]
	\caption{Approximate eigenspaces factorization.}
	\label{alg:fpca}
	\begin{algorithmic}
		\item[] {\bfseries Input:} The symmetric $\mathbf{S}$ or general $\mathbf{C}$, the size of the approximation $g$ or $m$, the update rule for the eigenvalues in \{`original', `update'\} and the stopping criterion $\epsilon$ (default taken to be $\epsilon = 10^{-2}$).
		
		\item[] {\bfseries Output:} The linear transformation $\mathbf{\bar{U}}$ and spectrum $\mathbf{\bar{s}}$ or linear transformation $\mathbf{\bar{T}}$ and spectrum $\mathbf{\bar{c}}$, the approximate solutions to \eqref{eq:symmetricproblem} or \eqref{eq:unsymmetricproblem}, respectively.\\
		
		\item[] {\bfseries Setup:} $\mathbf{G}_{i_k j_k} \! = \! \mathbf{I}_{n \times n},\ k\! =\! 1,\dots,g$ and compute all scores $\mathscr{A}_{ij}$ from to \eqref{eq:scriptA} or $\mathbf{T}_{i_k j_k} \! = \! \mathbf{I}_{n \times n},\ k\! =\! 1,\dots,m$ and compute all scores $\mathscr{C}_{ij}$ from to \eqref{eq:optC}; if the update rule for the spectrum is `update' and the true spectrum is not available then $\mathbf{\bar{s}} = \text{diag}(\mathbf{S})$ or, $\mathbf{\bar{c}} = \text{diag}(\mathbf{C})$, respectively.\\
		\item[]
		
		\item[] {\bfseries Initialization:} \textbf{for} $k=g$ \textbf{down to} 1 initialize each G-transform $\mathbf{G}_{i_k j_k}$ using Theorem 1 in the symmetric case or \textbf{for} $k=1$ \textbf{to} $g$ initialize each T-transform $\mathbf{T}_{i_k j_k}$ using Theorem 3 in the general case.
		
		{\bfseries Iterations:}
		
		$\bullet$ \textbf{for} $k = 1$ \textbf{to} $g$ update each G-transform $\mathbf{G}_{i_k j_k}$ (with all others fixed) according to Theorem 2 or \textbf{for} $k = 1$ \textbf{to} $m$ update each T-transform $\mathbf{T}_{i_k j_k}$ (with all others fixed) according to Theorem 4.
		
		$\bullet$ if rule is `update' calculate the new spectrum $\mathbf{\bar{s}}$ according to Lemma 1 or calculate the spectrum $\mathbf{\bar{c}}$ according to Lemma 2.
		
		$\bullet$ $i \leftarrow i+1$ and $\epsilon_i = \| \mathbf{S} - \mathbf{\bar{U}}\text{diag}(\mathbf{\bar{s}})\mathbf{\bar{U}}^T \|_F^2 $ or $\epsilon_i = \| \mathbf{C} - \mathbf{\bar{T}}\text{diag}(\mathbf{\bar{c}})\mathbf{\bar{T}}^{-1} \|_F^2 $.
		$ |\epsilon_{i-1} - \epsilon_i| < \epsilon$, if $i > 1$.
	\end{algorithmic}
\end{algorithm}
\begin{rem}\textbf{(Using T-transforms for the symmetric case)} The ideas of this section can also be applied to the symmetric case. Consider an approximation analogous to \eqref{eq:approximations4} based on T-transforms as
    \begin{equation}
    \mathbf{\bar{\bar{S}}} = \left( \prod_{k=1}^m \mathbf{T}_{i_k j_k} \right) \left( \prod_{k=m}^{1} \mathbf{T}_{i_k j_k}^T \right) = \mathbf{\bar{T}} \mathbf{\bar{T}}^T.
    \label{eq:approximations4-second}
    \end{equation}
    If the factorization in \eqref{eq:approximations4} is based on the eigendecomposition of symmetric matrices, this one is similar to a Cholesky factorization. With this structure we reach minimization problems similar to \eqref{eq:unsymmetricWithT} that have similar solutions to \eqref{eq:solutionsT} following the same development as in \eqref{eq:developmentforT}, for brevity we omit these formulas. The obvious disadvantage of this approach is that we do not explicitly preserve the eigen-information of the original matrix: we lose control over the spectrum of the approximation and the T-transform do not approximate explicitly the eigenspace. Still, there are some advantages: i) T-transforms (2 operations per degree of freedom) are more efficient than G-transforms (6 operations per degree of freedom) and therefore we expect to get better approximation accuracy for the same numerical complexity or vice-versa, lower computational complexity for the same representation accuracy; and ii) direct and inverse matrix-vector multiplications with $\mathbf{\bar{\bar{S}}}$ are still efficient, e.g., take at most $4g$ operations instead of $12g + n$ with $\mathbf{\bar{S}}$.
    
    An alternative construction is to keep the eigendecomposition but use real-valued approximate eigenvalues and use T-transforms to approximate the orthonormal eigenspace
    \begin{equation}
    \mathbf{\bar{\bar{S}}}  = \left( \prod_{k=1}^m \mathbf{T}_{i_k j_k}  \right) \text{diag}(\mathbf{\bar{s}}) \left( \prod_{k=m}^{1} \mathbf{T}_{i_k j_k}^{-1} \right)  = \mathbf{\bar{T}} \text{diag}(\mathbf{\bar{s}}) \mathbf{\bar{T}}^{-1}.
    \label{eq:approximations4-third}
    \end{equation}
    We lose the orthogonality of the eigenspace but we expect this representation to be more accurate just because, as already discussed, T-transforms have better numerical properties as compared to G-transforms. As every matrix in \eqref{eq:localstructure} can be written as four transformations from \eqref{eq:localstructure2} (two shears and two scalings by \citep{Lifting2}), we can use the approximation of $\mathbf{S}$ from \eqref{eq:approximations4} as an initialization to the factorization in \eqref{eq:approximations4-third} with $m = 4g$.
\end{rem}

\begin{rem}\textbf{(An approximate Schur decomposition for $\mathbf{S}$)} Notice that in \eqref{eq:symmetricproblem}, instead of the diagonal containing $\mathbf{\bar{s}}$ we can use for example an upper (or lower) triangular matrix which is also sparse, say $O(g)$ off-diagonal elements. The computational complexity of using such an approximation (directly or inversely) would still be $O(g)$ while we expect the approximation accuracy to be better, lower overall $\| \mathbf{S} -\mathbf{\bar{S}} \|_F^2$ due to the extra degrees in freedom in the new triangular factor. This factorization would be similar to the Schur decomposition $\mathbf{C} = \mathbf{V J V}^{-1}$, where $\mathbf{J}$ is upper triangular and $\mathbf{V}$ is orthonormal (but all real-valued). This is in contrast with the eigenvalues decomposition of $\mathbf{C}$ which is done over the complex values, in general.
\end{rem}

\subsection{The proposed algorithm}
\begin{figure*}[th]
	\centering
	\includegraphics[trim = 25 7 95 8, clip, width=0.9\textwidth]{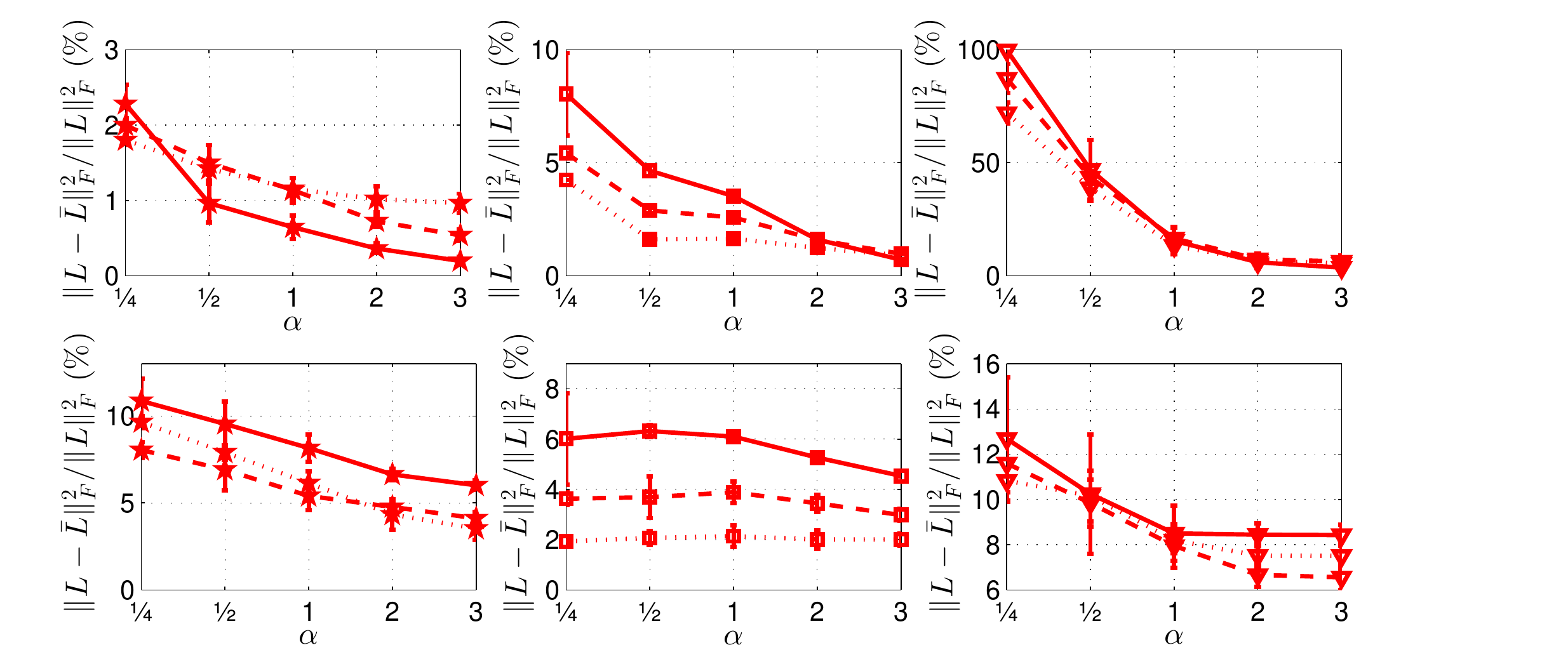}
	\caption{Approximation accuracy (mean and std) for Laplacians of randomly generated graphs as a function of the number of transformations $g$ going as $\alpha n \log_2 n $. All graphs are randomly generated using the default settings of the GSP box: community graphs (left), Erdos-Renyi random graphs with the probability of a connection between two nodes $p = 0.3$ (center) and a sensor graphs (right). Given $n$ the number of nodes in the graph, results are shown for $n = 512$ (dotted), $n=256$ (dashed) and $n=128$ (solid) and all methods update also the spectrum of the estimation. Top row shows undirected graphs while bottom row show directed graph (created from undirected graphs, direction of the edge between the nodes is decided randomly with probability 0.5) results. Results are averaged over 100 realizations.}
	\label{fig:figure2}
\end{figure*}
\begin{figure*}[!h]
	\centering
	\begin{minipage}[t]{0.32\textwidth}
		\includegraphics[trim = 18 5 33 15, clip, width=0.9\textwidth]{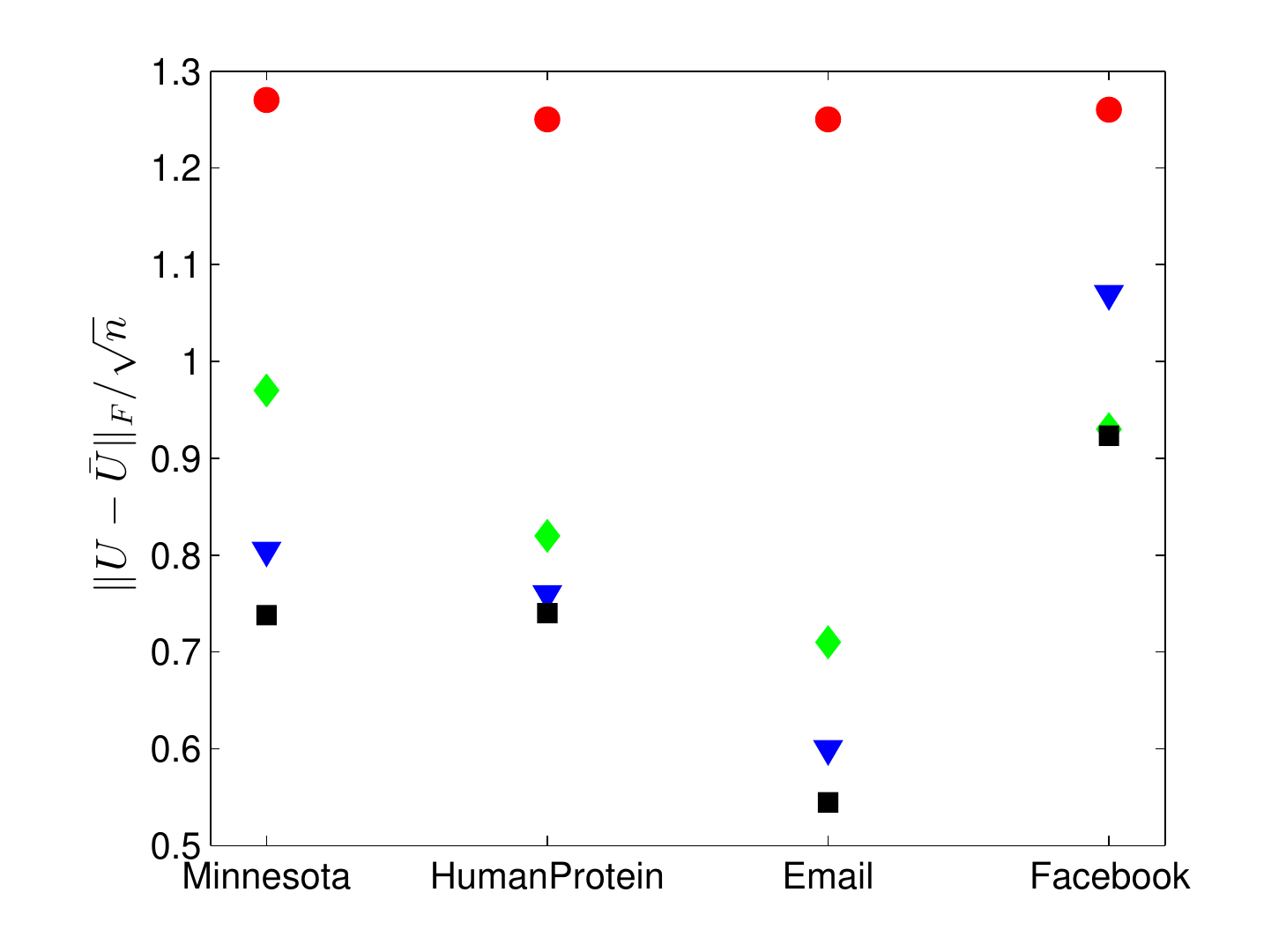}
		\caption{Comparison of the proposed approach (black squares) against previously proposed methods from the literature: Jacobi (red circles) from \citep{lemagoarou:hal-01416110}, greedy Givens (green diamonds) from \citep{Kondor2014MMF} and L1 (blue triangles) from \citep{Frerix2019ApproximatingOM}. The comparison is based on Fig. 6 from \citep{Frerix2019ApproximatingOM} and we keep their measure of accuracy between $\mathbf{U}$ and $\mathbf{\bar{U}}$.}
		\label{fig:figure3}
	\end{minipage}
	\hfill
	\begin{minipage}[t]{0.32\textwidth}
		\includegraphics[trim = 16 5 30 15, clip, width=0.9\textwidth]{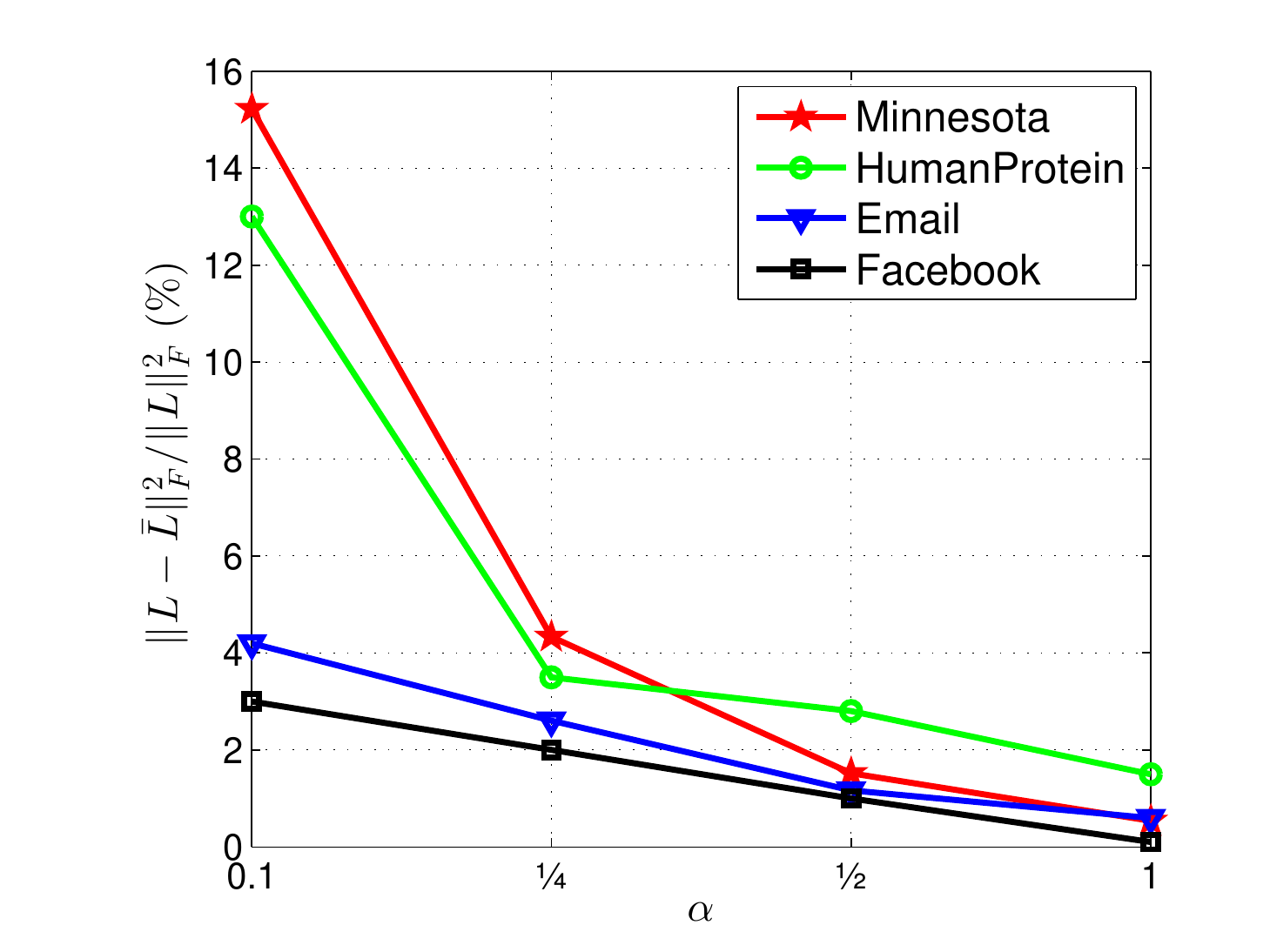}
		\caption{For the proposed method, we show average estimation accuracy for the overall Laplacian $\mathbf{L}$, not just the eigenspace $\mathbf{U}$, for the graphs in Figure \ref{fig:figure3} as a function of the number of G-transforms $g$ going as $\alpha n \log_2 n $. In all cases the proposed method also updates the spectrum of the approximation, i.e., $\mathbf{\bar{L}} = \mathbf{\bar{U}} \text{diag}(\bar{\lambda}) \mathbf{\bar{U}}^T$. The initial estimated eigenvalues are assumed to be the diagonal of $\mathbf{L}$.}
		\label{fig:figure4}
	\end{minipage}
	\hfill
	\begin{minipage}[t]{0.32\textwidth}
		\includegraphics[trim = 18 5 25 15, clip, width=0.9\textwidth]{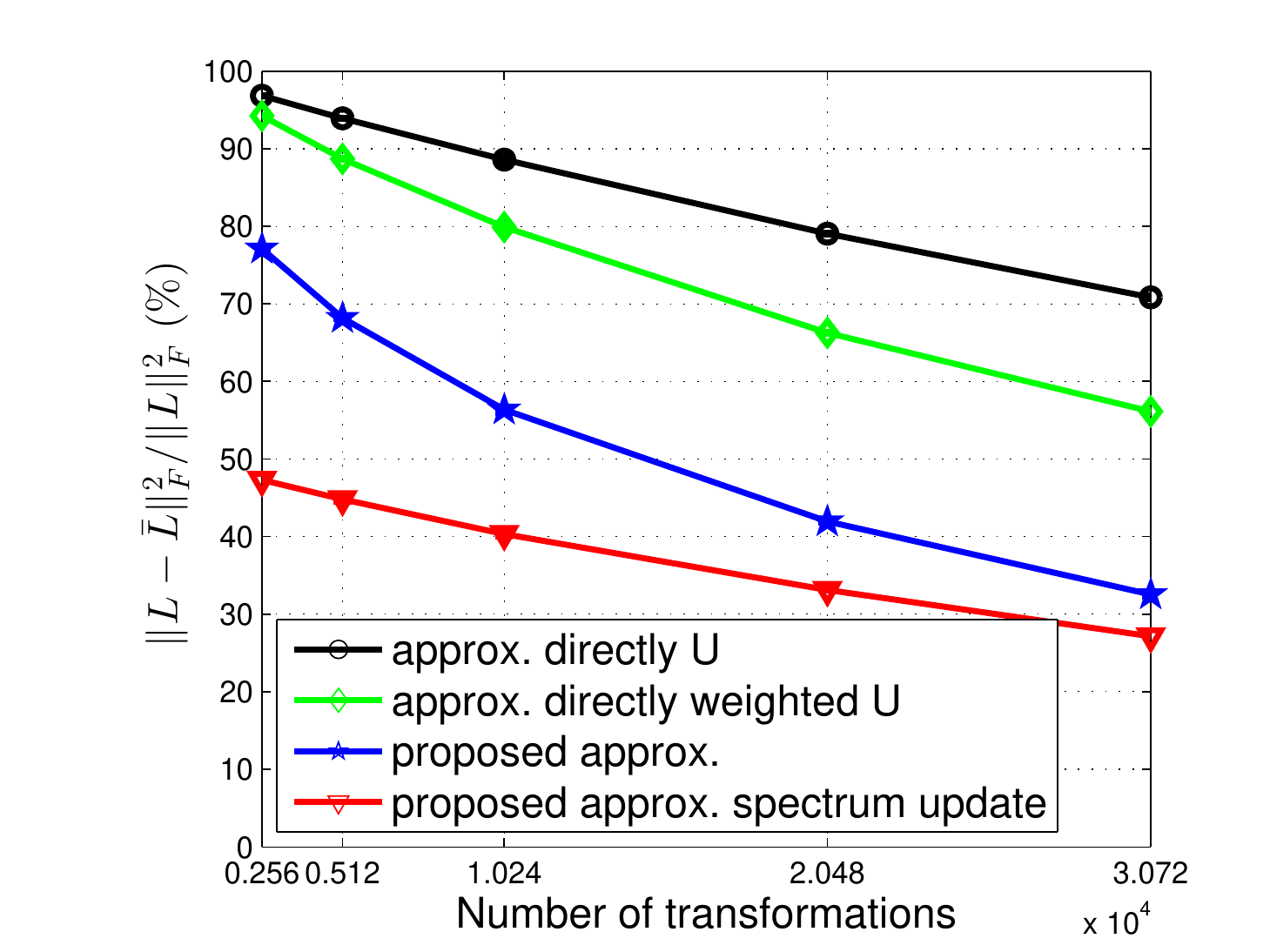}
		\caption{Given random undirected Erdos-Renyi graphs on size $n = 1024$ with Laplacian $\mathbf{L} = \mathbf{U}\text{diag}(\mathbf{\lambda}) \mathbf{U}^T$ we show the average approximation accuracy when we are approximating $\mathbf{U}$ directly, i.e., the eigendecomposition is explicitly given, or we are approximating $\mathbf{L}$ directly. When the eigendecomposition is available we approximate $\mathbf{U}$ and $\mathbf{U} \sqrt{\mathbf{\text{diag}(\lambda)}}$ (a weighted eigenspace) using \citep{FastPCA}.}
		\label{fig:figure5}
	\end{minipage}
\end{figure*}

\footnotetext{{https://epfl-lts2.github.io/gspbox-html}}
Given that we now have optimal ways to initialize (Theorems 1 and 3) and update G and T transforms (Theorems 2 and 4), we are ready the describe the full proposed procedure, in Algorithm 1. For brevity, we describe a single procedure for both the symmetric and general matrices. Whenever we allow for spectrum updates of $\mathbf{\bar{s}}$ or $\mathbf{\bar{c}}$ we explicitly mention so.

As previously discussed, every step of the algorithm is locally optimal and can only decrease the objective functions we consider. Thus, convergence to a stationary point is guaranteed. Compared to previous methods, there are two characteristics that we would like to highlight at this point: i) in the symmetric case, we use simultaneously both rotations and reflections to construct our approximation; and ii) for each sub-problem we define we can provide a closed-form solution based either on singular and eigenvalue decompositions or least squares. Also, because the proposed method relies on the calculation of scores that span indices $i$ and $j$ it is naturally parallelizable and is amenable to randomized linear algebra techniques.

Proofs of the lemmas and theorems from this section are collected in the supplementary materials.

\section{Experimental results}

In this section, we describe numerical experimental results with the proposed algorithms and compare them to previous work. Following previous methods \citep{lemagoarou:hal-01416110, Frerix2019ApproximatingOM}, we measure the quality of the approximation using the Frobenius norm objective functions of the optimization problems we consider. Source code for Algorithm 1 is available online\footnote{https://github.com/cristian-rusu-research/FAST-EIGENSPACE-APPROXIMATIONS}.
In all the experiments we use `update' for the spectrum estimation and only polishing in the iterative part of the algorithm (keep indices found in the initialization fixed and optimize only the values of the transforms - monotonicity in the objective function is still preserved).

We show an application to the calculation of the fast graph Fourier transforms. Given a graph with $n$ vertices we compute its $n \times n$ Laplacian $\mathbf{L} = \mathbf{D} - \mathbf{A}$ where $\mathbf{D}$ is the diagonal degree matrix and $\mathbf{A}$ is the $n \times n$ adjacency matrix, i.e., $A_{ij} = 1$ if a directed edge exists between nodes $i$ and $j$ and $A_{ij} = 0$ otherwise. Given the eigenvalue decomposition of the Laplacian, $\mathbf{L} = \mathbf{U \Lambda U}^{-1}$ we call $\mathbf{U}$ the graph Fourier transformation of the graph. Our goal is now to build approximations of $\mathbf{U}$ which enjoy the same numerical complexity as the classic time-domain Fourier transformation, i.e., $O(n \log n)$. We distinguish between undirected and directed graphs. With undirected graphs, the adjacency matrix $\mathbf{A}$ is symmetric and therefore the Laplacian is symmetric positive semidefinite allowing for an eigendecomposition with an orthonormal basis $\mathbf{U}$ as $\mathbf{L} = \mathbf{U \Lambda U}^{T}$. In this case, we use G-transforms to approximate the eigenspace. For directed graphs, the Laplacian does not have an orthonormal structure and therefore we use the more general T-transforms in the factorization. In Figure \ref{fig:figure2} we show approximation results for different types of graphs of different sizes (number of vertices $n$).

In Figure \ref{fig:figure3} we compared the proposed method against the previous state of the art for the computation of fast graph Fourier transforms. The results are shown for four graphs: Minnesota graph from \citep{pygsp} with $n = 2642$ and 3304 edges, HumanProtein graph from \citep{Rual2005} with $n=3133$ and 6726 edges, Email graph from \citep{Guimera2004} with $n = 1133$ and 5451 edges and the Facebook graph from \citep{NIPS2012_4532} with $n = 2888$ and 2981 edges. Our proposed method performs best in all these situations.

In this paper, we assume we do not have information about the eigenspace of $\mathbf{L}$, but have access to $\mathbf{L}$ itself. Previous work \citep{FastPCA, Frerix2019ApproximatingOM} considered the possibility of performing first the eigendecomposition and then working directly with the eigenspace $\mathbf{U}$. For the same graphs, we show in Figure \ref{fig:figure4} the evolution of the accuracy of the overall Laplacian $\mathbf{L}$ as a function of the number of basic transformations in their factorization. We emphasize again that this metric is different from the one used in Figure \ref{fig:figure3} on the accuracy of the eigenspace $\mathbf{U}$.

In Figure \ref{fig:figure5} we analyze the approximation accuracy of estimating the Laplacian $\mathbf{L}$ for random undirected Erdos-Renyi graphs with $n = 1024$ and compare it against the approach in \citep{FastPCA} which needs directly the eigenspace $\mathbf{U}$. We show that our proposed method, especially with the updated spectrum, is the most appropriate to build an accurate approximation of $\mathbf{L}$.

The supplementary materials have further numerical experiments, comparisons, and measurements of the running time of the fast transformations (not just number of operations).

\section{Conclusions}

In this paper, we have described algorithms for the efficient approximate computation of eigenspaces which can be efficiently manipulated. We show an application to the computation of the fast graph Fourier transform (both for directed and undirected graphs) and compare against previous approaches from the literature, which we outperform. An open problem, that we cannot address at this time, is the setup of an appropriate theoretical framework where the proposed factorizations and algorithms can be analyzed and their limitations understood.

\section*{Acknowledgments}
This material is based upon work supported by the Center for Brains, Minds and Machines (CBMM), funded by NSF STC award CCF-1231216, and the Italian Institute of Technology. Part of this work has been carried out at the Machine Learning Genoa (MaLGa) center, Università di Genova (IT) L. Rosasco acknowledges the financial support of the European Research Council (grant SLING 819789), the AFOSR projects FA9550-17-1-0390 and BAA-AFRL-AFOSR-2016-0007 (European Office of Aerospace Research and Development), and the EU H2020-MSCA-RISE project NoMADS - DLV-777826. C. Rusu is supported by
the Romanian Ministry of Education and Research, CNCS-UEFISCDI,
project number PN-III-P1-1.1-TE-2019-1843, within PNCDI III.

\newpage

\section*{Supplementary materials}

\subsection*{Eigenvalues of $2 \times 2$ symmetric matrices}

Given the symmetric $2 \times 2$ matrix $\mathbf{S}_{ \{i,j\} } = \begin{bmatrix}
S_{ii} & S_{ij} \\
S_{ij} & S_{jj}
\end{bmatrix}$ its two eigenvalues are given by
\begin{equation}
\lambda_{1,2} = \frac{1}{2} \left( S_{ii}+S_{jj} \pm \sqrt{ (S_{ii}-S_{jj})^2 + 4S_{ij}^2  } \right).
\label{eq:eigenvalues}
\end{equation}

\subsection*{Proof of Lemma 1}

The result follows directly by using the invariance of the Frobenius norm under orthogonal transformations:
\begin{equation}
\| \mathbf{S}  - \mathbf{\bar{U}} \text{diag}(\mathbf{\bar{s}}) \mathbf{\bar{U}}^T \|_F^2 = \| \mathbf{\bar{U}}^T \mathbf{S} \mathbf{\bar{U}} -  \text{diag}(\mathbf{\bar{s}})  \|_F^2.
\end{equation}
Then, because the Frobenius norm is entry-wise we have the minimizer $\mathbf{\bar{s}}^\star =  \text{diag}(\mathbf{\bar{U}}^T \mathbf{S} \mathbf{\bar{U}})$.

\subsection*{Proof of Theorem 1}

Given that we have initial values for all components $\mathbf{G}_{i_t j_t}$ for $t = g,\dots,k+1$ (while $\mathbf{G}_{i_t j_t} = \mathbf{I}_n$ for $t=1,\dots,k-1$) and we now want to also initialize the $k^\text{th}$ component such that we minimize objective function of \eqref{eq:symmetricproblem}. That expression can be written as
\begin{equation}
\begin{aligned}
\| &\mathbf{S} - \mathbf{\bar{S}} \|_F^2 = \| \mathbf{S} - \mathbf{\bar{U}} \text{diag}(\mathbf{\bar{s}}) \mathbf{\bar{U}}^T \|_F^2 \\
= & \Bigg\| \mathbf{S} - \left. \prod_{t=k}^g \mathbf{G}_{i_t j_t} \text{diag}(\mathbf{\bar{s}}) \prod_{t=g}^{k} \mathbf{G}_{i_t j_t}^T \right\|_F^2 \\
= & \left\| \prod_{t=g}^{k+1} \mathbf{G}_{i_t j_t}^T \mathbf{S} \! \! \! \prod_{t=k+1}^{g} \mathbf{G}_{i_t j_t}^T \! - \mathbf{G}_{i_k j_k} \text{diag}(\mathbf{\bar{s}}) \mathbf{G}_{i_k j_k}^T \right\|_F^2 \\
= & \|  \mathbf{S}^{(k)} - \mathbf{G}_{i_k j_k} \text{diag}(\mathbf{\bar{s}}) \mathbf{G}_{i_k j_k}^T \|_F^2 \\
= & \| \mathbf{S}^{(k)} \|_F^2 + \| \mathbf{\bar{s}} \|_2^2 - 2\text{tr}(\mathbf{Z}^{(k)}) - 2\mathscr{A}_{i_k j_k} \\
= & \| \mathbf{s} \|_2^2 + \| \mathbf{\bar{s}} \|_2^2 - 2\text{tr}(\mathbf{Z}^{(k)}) - 2\mathscr{A}_{i_k j_k},
\end{aligned}
\label{eq:developmentinitialization}
\end{equation}
where we have defined the cost
\begin{equation}
\mathscr{A}_{i_k j_k} \! = \! \text{tr}(\mathbf{\tilde{G}}_{k} \mathbf{S}^{(k)}_{ \{i_k, j_k \} } \! \mathbf{\tilde{G}}_{k}^T \text{diag}(\mathbf{\bar{s}}_{ \{i_k, j_k \}})  ) - Z_{i_k i_k}^{(k)} \! -\! Z_{j_k j_k}^{(k)}.
\label{eq:theC}
\end{equation}
For convenience, we defined the $2 \times 2$ symmetric matrices
\begin{equation}
\mathbf{S}^{(k)}_{\{i_k, j_k\} } = \begin{bmatrix}
S^{(k)}_{i_k i_k} & S^{(k)}_{i_k j_k} \\
S^{(k)}_{j_k i_k} & S^{(k)}_{j_k j_k}
\end{bmatrix},
\end{equation}
\begin{equation}
\text{diag}(\mathbf{\bar{s}}_{\{i_k, j_k \} }) = \text{diag}(\begin{bmatrix}
\bar{s}_{i_k i_k} & \bar{s}_{j_k j_k}
\end{bmatrix}),
\end{equation}
and $\mathbf{Z}^{(k)} = \text{diag}(\mathbf{\bar{s}}) \mathbf{S}^{(k)} \in \mathbb{R}^{ n \times n}$. In the development of \eqref{eq:developmentinitialization} we have used the trace definition of the Frobenius norm $\|\mathbf{X}\|_F^2 = \text{tr}(\mathbf{X}^T \mathbf{X})$, the fact that the Frobenius norm is invariant to orthonormal transformation (in particular G-transformation) $\|\mathbf{G}_{i_k j_k} \mathbf{X}\|_F^2 = \|\mathbf{G}_{i_k j_k}^T \mathbf{X}\|_F^2 = \|\mathbf{X}\|_F^2$ and that $\mathbf{G}_{i_k j_k}$ operates only on rows and columns $i_k$ and $j_k$.\\
The problem of maximizing the quantity in \eqref{eq:theC} is known as the two-side orthonormal Procrustes problem \citep{Schonemann1968} whose solution in our case is given by
\begin{equation}
\mathbf{\tilde{G}}_{k} = \mathbf{V}_k^T \text{ where } \mathbf{S}^{(k)}_{ \{i_k, j_k \} } = \mathbf{V}_k \mathbf{D}_k \mathbf{V}_k^T.
\label{eq:optimalinitializationgikjk}
\end{equation}
We will use that $\mathbf{V}_k\mathbf{V}_k^T = \mathbf{V}_k^T \mathbf{V}_k = \mathbf{I}$. We assume that in the eigenvalue decomposition of $\mathbf{S}^{(k)}_{i_k j_k}$ where the diagonal matrix $\mathbf{D}_k = \text{diag}(\mathbf{d}_k)$ contains the eigenvalues in algebraic descending order. The same ordering is also assumed in $\mathbf{\bar{s}}$. Therefore, by the rearrangement inequality, see Section 10.2,~Theorem 368 of \citep{Hardy}, and with $j_k > i_k$ the trace quantity is maximized and it reduces to
\begin{equation}
\begin{aligned}
\text{tr}(\mathbf{V}_k^T \mathbf{S}^{(k)}_{ \{i_k, j_k\} } & \mathbf{V} _k \text{diag}(\mathbf{\bar{s}}_{ \{i_k, j_k \}})  ) \\
= & \text{tr}(\! \mathbf{V}_k^T \mathbf{V}_k \mathbf{D}_k \! \mathbf{V}_k^T \mathbf{V}_k \text{diag}(\mathbf{\bar{s}}_{ \{i_k, j_k \} }  )  ) \\
= &  \text{tr}(\mathbf{D}_k \text{diag}(\mathbf{\bar{s}}_{ \{i_k, j_k\}} \! ) \! ) \\
= & \text{tr}(\text{diag}(\mathbf{d}_k) \text{diag}(\mathbf{\bar{s}}_{ \{i_k, j_k\} }  )) \\
= &  \begin{bmatrix}
\bar{s}_{i_k i_k} & \bar{s}_{j_k j_k}
\end{bmatrix}^T \mathbf{d}_k.
\end{aligned}
\end{equation}
Therefore, the overall cost \eqref{eq:theC} reduces to
\begin{equation}
\begin{aligned}
& \mathscr{A}_{i_k j_k} = \begin{bmatrix}
\bar{s}_{i_k i_k} & \bar{s}_{j_k j_k}
\end{bmatrix}^T \mathbf{d}_k - Z_{i_k i_k}^{(k)} - Z_{j_k j_k}^{(k)} \\
= & \begin{bmatrix}
\bar{s}_{i_k i_k} & \bar{s}_{j_k j_k}
\end{bmatrix}^T \left( \mathbf{d}_k - \begin{bmatrix}
S^{(k)}_{i_k i_k}  \\
S^{(k)}_{j_k j_k}
\end{bmatrix}  \right)\\
= & \begin{bmatrix}
\bar{s}_{i_k i_k} & \bar{s}_{j_k j_k}
\end{bmatrix}^T \! \begin{bmatrix}
-\gamma_{ i_k j_k }  \\
\gamma_{ i_k j_k }
\end{bmatrix}  \\
=& \gamma_{ i_k j_k } (\bar{s}_{j_k j_k} - \bar{s}_{i_k i_k}),
\end{aligned}
\label{eq:thecikjk}
\end{equation}
where we have denoted
\begin{equation}
\gamma_{ i_k j_k }  =  \frac{S^{(k)}_{i_k i_k }  - S^{(k)}_{ j_k j_k }}{2} \! \! \left( 1 + \sqrt{  1 + \left( \frac{2S^{(k)}_{i_k j_k }}{S^{(k)}_{ i_k i_k }-S^{(k)}_{ j_k j_k }} \right)^2} \right),
\end{equation}
and we noticed that $Z_{i_k i_k}^{(k)} = \bar{s}_{i_k i_k} S^{(k)}_{i_k i_k} $ and $Z_{j_k j_k}^{(k)} = \bar{s}_{j_k j_k} S^{(k)}_{j_k j_k}$. The eigenvalues of $\mathbf{S}^{(k)}_{ \{i_k i_k \}}$ in $\mathbf{d}_k$ are computed by the formulas in \eqref{eq:eigenvalues}. Therefore, the minimizer of \eqref{eq:developmentinitialization} is
\begin{equation}
(i_k^\star, j_k^\star) = \underset{(i,j),\ j > i}{\arg \max} \ \mathscr{A}_{i j} \text{ and } \mathbf{\tilde{G}}^\star_{k} = \mathbf{V}_k^T.
\end{equation}

\subsection*{Proof of Theorem 2}

With the definitions \eqref{eq:ak} and \eqref{eq:bk}, the objective function \eqref{eq:symmetricWithGivens} can be expressed to
\begin{equation}
\begin{aligned}
\|  \mathbf{A}^{(k)} - & \mathbf{G}_{i_k j_k} \mathbf{B}^{(k)} \mathbf{G}_{i_k j_k}^T  \|_F^2 \\
= &  \| \mathbf{A}^{(k)} \|_F^2 + \| \mathbf{B}^{(k)} \|_F^2 - 2\text{tr}(\mathbf{Z}^{(k)}) -  2\mathscr{B}_{i_k j_k} \\
= & \| \mathbf{s} \|_2^2 + \| \mathbf{\bar{s}} \|_2^2 - 2\text{tr}(\mathbf{Z}^{(k)}) - 2\mathscr{B}_{i_k j_k},
\end{aligned}
\label{eq:iterativemorethanJacobi}
\end{equation}
with the cost that is
\begin{equation}
\begin{aligned}
\! \! \! \! \mathscr{B}_{i_k j_k} & = \! - 2(Z_{i_k i_k}^{(k)} \!  + Z_{j_k j_k}^{(k)}) \!  + V_{i_k i_k}^{(k)} \! + V_{j_k j_k}^{(k)} \! + 2V_{i_k j_k}^{(k)}\\
& + \text{tr}(\mathbf{A}_{\{i_k, j_k\}}^{(k)} \mathbf{\tilde{G}}_{k}\mathbf{B}_{\{i_k, j_k\}}^{(k)}\mathbf{\tilde{G}}_{k}^T) \\
& + 2\text{tr}(\mathbf{\tilde{G}}_{k}\mathbf{B}_{[i_k,j_k]}^{(k)}{\mathbf{A}_{[i_k,j_k]}^{(k)^T}}) ,
\end{aligned}
\label{eq:nasty}
\end{equation}
where we have defined $\mathbf{A}^{(k)}_{ \{i_k,j_k\} } = \begin{bmatrix}
A_{i_k i_k}^{(k)} & A_{i_k j_k}^{(k)}\\
A_{j_k i_k}^{(k)} & A_{j_k j_k}^{(k)}
\end{bmatrix}$, $\mathbf{B}^{(k)}_{ \{i_k,j_k\} } = \begin{bmatrix}
B_{i_k i_k}^{(k)} & B_{i_k j_k}^{(k)}\\
B_{j_k i_k}^{(k)} & B_{j_k j_k}^{(k)}
\end{bmatrix}$, $\mathbf{A}_{[i_k,j_k]}^{(k)}$ and $\mathbf{B}_{[i_k,j_k]}^{(k)}$ are both matrices of size $2 \times (n-2)$ composed of only rows $i_k$ and $j_k$, but with the columns $i_k$ and $j_k$ eliminated from both, from $\mathbf{A}^{(k)}$ and $\mathbf{B}^{(k)}$ respectively. Finally, we have used
\begin{equation}
\mathbf{Z}^{(k)} = \mathbf{A}^{(k)}\mathbf{B}^{(k)},\ \mathbf{V}^{(k)} = \mathbf{A}^{(k)}\odot \mathbf{B}^{(k)},
\end{equation}
the operation $\odot$ denotes the entry-wise matrix-matrix product.\\
The quantity in \eqref{eq:nasty} seems difficult to minimize, in the sense that a solution based on an eigenvalue decomposition, such as in \eqref{eq:optimalinitializationgikjk}, does not seems possible. This is because the cost contains a term (the first trace, similar to the one in \eqref{eq:theC}) whose maximum is given as the solution to the two-side orthonormal Procrustes problem \citep{Schonemann1968} by eigenvalue decompositions of $\mathbf{A}^{(k)}_{ \{i_k,j_k\} } $ and $\mathbf{B}^{(k)}_{ \{i_k,j_k\} } $ but also another term (the second trace) whose maximum is given as the solution to the one-sided orthogonal Procrustes problem by the singular value decomposition of another quantity, $\mathbf{B}_{[i_k,j_k]}^{(k)}{\mathbf{A}_{[i_k,j_k]}^{(k)^T}}$. The computational simplification in \eqref{eq:theC} is possible because one of the matrices is diagonal and therefore the second trace term does not appear in the cost. It is for this reason that we have to analyze separately the initialization and iteratively procedures.\\
To find the minimizer of \eqref{eq:symmetricWithGivens}, both in terms of the indices and the values of the orthonormal $\mathbf{G}_{i_k j_k}$, take the following
\begin{equation}
\begin{aligned}
\| & \mathbf{A}^{(k)}  - \mathbf{G}_{i_k j_k} \mathbf{B}^{(k)}  \mathbf{G}_{i_k j_k}^T \|_F^2 \\
= & \| \mathbf{A}^{(k)} \mathbf{G}_{i_k j_k} - \mathbf{G}_{i_k j_k} \mathbf{B}^{(k)}   \|_F^2 \\
= &  \| \text{vec}(\mathbf{A}^{(k)} \mathbf{G}_{i_k j_k})  - \text{vec}(\mathbf{G}_{i_k j_k} \mathbf{B}^{(k)} )  \|_2^2 \\
= & \| (\mathbf{I} \otimes \mathbf{A}^{(k)})\text{vec}(\mathbf{G}_{i_k j_k}) - (\mathbf{B}^{(k)} \otimes \mathbf{I})\text{vec}(\mathbf{G}_{i_k j_k}) \|_2^2 \\
= & \| ( (\mathbf{I} \otimes \mathbf{A}^{(k)}) - (\mathbf{B}^{(k)} \otimes \mathbf{I}) )\text{vec}(\mathbf{G}_{i_k j_k}) \|_2^2 \\
= & \left\| \sum_{t \in \{1,\dots,n\}\backslash \{i_k, j_k\} } \! \! \! \! \! \! \left( \mathbf{e}_t \otimes \mathbf{A}^{(k)}_{:,t} - \mathbf{B}^{(k)}_{:, t} \otimes \mathbf{e}_t \right) \! + \! \mathbf{P} \begin{bmatrix} c_{i_k j_k} \\ s_{i_k j_k} \end{bmatrix} \! \right\|_2^2 \\
= & \left\| \mathbf{w} + \mathbf{P}\begin{bmatrix} c_{i_k j_k} \\ s_{i_k j_k} \end{bmatrix} \right\|_2^2,
\end{aligned}
\label{eq:morethanJacobi}
\end{equation}
where $(\mathbf{A}^{(k)})_{:, t}$ and $(\mathbf{B}^{(k)})_{:, t}$ are the $t^\text{th}$ columns of $\mathbf{A}^{(k)}$ and $\mathbf{B}^{(k)}$, respectively. We have introduced the matrix
\begin{equation}
\! \mathbf{P} \! \in \!  \left\{ \!  \begin{bmatrix}
\mathbf{p}_1 \! + \! \mathbf{p}_2 & \! \mathbf{p}_3 \! - \! \mathbf{p}_4
\end{bmatrix}\!,\!
\begin{bmatrix}
\mathbf{p}_1 \! - \! \mathbf{p}_2 & \! \mathbf{p}_3 \! + \! \mathbf{p}_4
\end{bmatrix} \! \right\} \! \! \in \! \mathbb{R}^{n^2 \times 2},
\label{eq:theZ}
\end{equation}
with $\mathbf{p}_1 = \mathbf{e}_{i_k} \otimes (\mathbf{A}^{(k)})_{:, i_k} - (\mathbf{B}^{(k)})_{:, i_k} \otimes \mathbf{e}_{i_k},\ \mathbf{p}_2 = \mathbf{e}_{j_k} \otimes (\mathbf{A}^{(k)})_{:, j_k} - (\mathbf{B}^{(k)})_{:, j_k} \otimes \mathbf{e}_{j_k},\ \mathbf{p}_3 = \mathbf{e}_{j_k} \otimes (\mathbf{A}^{(k)})_{:, i_k} - (\mathbf{B}^{(k)})_{:, j_k} \otimes \mathbf{e}_{i_k}$ and $\mathbf{p}_4 = \mathbf{e}_{i_k} \otimes (\mathbf{A}^{(k)})_{:, j_k} - (\mathbf{B}^{(k)})_{:, i_k} \otimes \mathbf{e}_{j_k}$ where $\{ \mathbf{e}_i \}_{i=1}^n$ are the standard basis vectors for $\mathbb{R}^n$ and $\otimes$ is the Kronecker product. For ease, we also define $\mathbf{P}^{(1)}$ and $\mathbf{P}^{(2)}$, the two options for $\mathbf{P}$. We have these two variants for $\mathbf{P}$ due to the dual structure of \eqref{eq:localstructure}. Unlike the previous section, where the dual structure enabled the initialization by an eigenvalue decomposition, here we actually need to solve two different, but related, problems.\\
In the development of \eqref{eq:morethanJacobi} we have used again the invariance of norms to orthonormal transformations and the fact that the Frobenius norm is element-wise $\|\mathbf{X}\|_F^2 = \| \text{vec}(\mathbf{X}) \|_F^2$ and that $\text{vec}(\mathbf{ABC}) = (\mathbf{C}^T \otimes \mathbf{A}) \text{vec}(\mathbf{B})$. The rest of the section is dedicated to finding the minimizer of \eqref{eq:morethanJacobi} in a numerically efficient manner.\\
Therefore, the minimization of \eqref{eq:morethanJacobi} is equivalent to a constrained least squares problem that can be solved efficiently using the singular value decomposition, see Chapter 12.1 of \citep{Golub1996} and \citep{SphereLS}. We have to solve the problem twice, for  $\mathbf{P}^{(1)}$ and $\mathbf{P}^{(2)}$ in \eqref{eq:theZ}, and keep the best result. The vector $\mathbf{w}$ and the matrix $\mathbf{P}$ are never explicitly constructed, but we build the products
\begin{equation}
\mathbf{R} = \mathbf{P}^T \mathbf{P} \in \mathbb{R}^{2 \times 2} \text{ and } \mathbf{g} = \mathbf{P}^T \mathbf{w} \in \mathbb{R}^{2 \times 1},
\end{equation}
For these, we have the explicit formulas
\begin{itemize}
    \item for $\mathbf{R}^{(1)} = {\mathbf{P}^{(1)}}^T \mathbf{P}^{(1)}$ we have: $R^{(1)}_{11} = W_{\text{A}i_k}^{(k)}+ W_{\text{A}j_k}^{(k)} +W_{\text{B}i_k}^{(k)}+ W_{\text{B}j_k}^{(k)}  - 2V_{i_k i_k}^{(k)} - 2V_{j_k j_k}^{(k)}- 4V_{i_k j_k}^{(k)}$, $R^{(1)}_{12} = R^{(1)}_{21} = 2(A_{i_k j_k}^{(k)} B_{i_k i_k}^{(k)} - A_{i_k i_k}^{(k)} B_{i_k j_k}^{(k)}  + A_{j_k j_k}^{(k)} B_{i_k j_k}^{(k)} - A_{i_k j_k} B_{j_k j_k} )$ and $R^{(1)}_{22} = W_{\text{A} i_k}^{(k)} + W_{\text{A}j_k}^{(k)} +  W_{\text{B} i_k}^{(k)} + W_{\text{B} j_k}^{(k)} - 2A_{i_k i_k}^{(k)} B_{j_k j_k}^{(k)}  - 2A_{j_k j_k}^{(k)} B_{i_k i_k}^{(k)} + 4V_{i_k j_k}^{(k)}$;
    \item for $\mathbf{g}^{(1)} = {\mathbf{P}^{(1)}}^T \mathbf{w}$ we have: $g^{(1)}_1 = 2(V_{i_k i_k}^{(k)} + V_{j_k j_k}^{(k)} + 2V_{i_k j_k}^{(k)} - Z_{i_k i_k}^{(k)} -Z_{j_k j_k}^{(k)} )$ and $g^{(1)}_2 = 2(A_{i_k j_k}^{(k)} B_{j_k j_k}^{(k)} + A_{i_k i_k}^{(k)} B_{i_k j_k}^{(k)} - A_{j_k j_k}^{(k)} B_{i_k j_k}^{(k)} - A_{i_k j_k}^{(k)} B_{i_k i_k}^{(k)}-Z_{i_k j_k}^{(k)}  + Z_{j_k i_k}^{(k)}  )$;
    \item for $\mathbf{R}^{(2)} = {\mathbf{P}^{(2)}}^T \mathbf{P}^{(2)}$ we have: $R^{(2)}_{11} = W_{\text{A}i_k}^{(k)}+ W_{\text{A}j_k}^{(k)} +W_{\text{B}i_k}^{(k)}+ W_{\text{B}j_k}^{(k)}  - 2V_{i_k i_k}^{(k)} - 2V_{j_k j_k}^{(k)}+ 4V_{i_k j_k}^{(k)}$, $R^{(2)}_{12} = R^{(2)}_{21} = 2(A_{i_k j_k }^{(k)} B_{j_k j_k}^{(k)} -A_{i_k i_k}^{(k)} B_{i_k j_k}^{(k)} + A_{j_k j_k}^{(k)} B_{i_k j_k}^{(k)} - A_{i_k j_k}^{(k)} B_{i_k i_k}^{(k)} )$ and $R^{(2)}_{22} = W_{\text{A} i_k}^{(k)}  + W_{\text{A} j_k}^{(k)}+ W_{\text{B} i_k}^{(k)} + W_{\text{B} j_k}^{(k)}   - 2A_{i_k i_k}^{(k)} B_{j_k j_k}^{(k)} - 2A_{j_k j_k}^{(k)} B_{i_k i_k}^{(k)} - 4V_{i_k j_k}^{(k)}$;
    \item for $\mathbf{g}^{(2)} = {\mathbf{P}^{(2)}}^T \mathbf{w}$ we have: $g^{(2)}_1 = 2(V_{i_k i_k}^{(k)} - V_{j_k j_k}^{(k)} - Z_{i_k i_k}^{(k)}  + Z_{j_k j_k}^{(k)} )$ and $g^{(2)}_2 = 2( A_{i_k j_k}^{(k)} B_{j_k j_k}^{(k)} + A_{i_k i_k}^{(k)} B_{i_k j_k}^{(k)} + A_{j_k j_k}^{(k)} B_{i_k j_k}^{(k)} + A_{i_k j_k}^{(k)} B_{i_k i_k}^{(k)} -Z_{i_k j_k}^{(k)} - Z_{j_k i_k}^{(k)})$.
\end{itemize}
Here we have also introduces the quantities:
\begin{equation}
W_{\text{A} i}^{(k)} \! = \! \| \mathbf{A}^{(k)}_{:,i} \|_2^2,\ W_{\text{B} i}^{(k)} \! = \! \| \mathbf{B}^{(k)}_{:,i} \|_2^2,\ i=1,\dots,n,
\end{equation}
that compute to the squared $\ell_2$ norms of the columns of $\mathbf{A}^{(k)}$ and $\mathbf{B}^{(k)}$, respectively. 
With this setup, the minimizer of \eqref{eq:morethanJacobi} with some fixed indices $i_k$ and $j_k$ is
\begin{equation}
\mathbf{x}^{(f_k)} = \begin{bmatrix} c_{i_k j_k}^\star \\ s_{i_k j_k}^\star \end{bmatrix} = -(\mathbf{R}^{(f_k)} + \lambda^{(f_k)} \mathbf{I}_2)^{-1} \mathbf{g}^{(f_k)},
\label{eq:bestortho}
\end{equation}
where $\lambda^{(f_k)}$ is chosen such that the solution $\mathbf{x}^{(f_k)}$ has unit $\ell_2$ norm according to \citep{SphereLS}, the regularization parameter is computed by
\begin{equation}
\lambda^{(f_k)} = \min\ \{ \lambda_i \},\text{ where } \mathbf{M}^{(f_k)} \mathbf{v}_i = \lambda_i \mathbf{N}^{(f_k)} \mathbf{v}_i,
\end{equation}
i.e., the $\lambda_i$ are the generalized real-valued eigenvalues of the $4 \times 4$ matrices $\mathbf{M} = \begin{bmatrix}
(\mathbf{R}^{(f_k)} )^2-\mathbf{g}^{(f_k)} (\mathbf{g}^{(f_k)})^T & \mathbf{0}_{2} \\
\mathbf{0}_{2} & \mathbf{I}_2
\end{bmatrix}$ and $\mathbf{N} = \begin{bmatrix}
2\mathbf{R}^{(f_k)} & -\mathbf{I}_2 \\
\mathbf{I}_2 & \mathbf{0}_{2}
\end{bmatrix}$. Therefore, using \eqref{eq:bestortho}, the overall minimizer of \eqref{eq:iterativemorethanJacobi} is found by searching for the indices
\begin{equation}
(f_k^\star, i_k^\star, j_k^\star) = \underset{f_k \in \{1,2\},\ j_k > i_k}{\arg \min} \mathscr{B}_{i_k j_k}^{(f_k)},
\end{equation}
where $\mathscr{B}_{i_k j_k}^{(f_k)} =  (\mathbf{x}^{(f_k)})^T \mathbf{R}^{(f_k)} \mathbf{x}^{(f_k)} + 2(\mathbf{x}^{(f_k)})^T \mathbf{g}^{(f_k)} + \| \mathbf{w} \|_2^2$, and then
\begin{equation}
\mathbf{\tilde{G}}^\star_{k} = \begin{cases}  \begin{bmatrix}
c_{i^\star_k j^\star_k}^\star & s_{i^\star_k j^\star_k}^\star\\
-s_{i^\star_k j^\star_k}^\star & c_{i^\star_k j^\star_k}^\star 
\end{bmatrix}, & \text{ if } f_k^\star = 1,  \\
\vspace*{-13pt}\\
\begin{bmatrix}
c_{i^\star_k j^\star_k}^\star & s_{i^\star_k j^\star_k}^\star\\
s_{i^\star_k j^\star_k}^\star & -c_{i^\star_k j^\star_k}^\star 
\end{bmatrix}, & \text{ if } f_k^\star=2.
\end{cases}
\end{equation}
The quantities $\mathbf{R}^{(f_k)}$, $\mathbf{g}^{(f_k)}$ are already computed and $\| \mathbf{w} \|_2^2 \! = \! \sum_{i \neq \{i_k, j_k\} } (W_{\text{A}i}^{(k)} + \! W_{\text{B}i}^{(k)}) \! - \! \sum_{i\neq \{i_k, j_k\} } \sum_{j\neq \{ i_k,j_k\} } 2 V_{i j}^{(k)}$.

\subsection*{Proof of Lemma 2}

The result follows by using the trace product formula:
\begin{equation}
\begin{aligned}
\| \mathbf{C}  - \mathbf{\bar{T}} \text{diag}(\mathbf{\bar{c}}) \mathbf{\bar{T}}^{-1} \|_F^2 = & \| \text{vec}(\mathbf{C}) \! - \! (\mathbf{\bar{T}}^{-T} \! \otimes \! \mathbf{\bar{T}} ) \text{diag}(\mathbf{\bar{c}}) \|_F^2 \\
= & \| \text{vec}(\mathbf{C}) -( \mathbf{\bar{T}}^{-T} * \mathbf{\bar{T}} ) \mathbf{\bar{c}} \|_F^2,
\end{aligned}
\end{equation}
Then, this is a simple least-squares problem in $\mathbf{\bar{c}}$ where $*$ is the Khatri-Rao product, the Kronecker products between the corresponding columns of $\mathbf{\bar{T}}^{-T}$ and $\mathbf{\bar{T}}$.

\subsection*{Proof of Theorem 3}

Consider the scenario where we have initial values for all components $\mathbf{T}_{i_t j_t}$ for $t = 1,\dots,k-1$ (while $\mathbf{T}_{i_t j_t} = \mathbf{I}_n$ for $t=k+1,\dots,m$) and we want to also initialize the $k^\text{th}$ component such that we minimize the quantity
\begin{equation}
\begin{aligned}
\Bigg\| & \mathbf{C} - \prod_{t=1}^k \mathbf{T}_{i_t j_t} \text{diag}(\mathbf{\bar{c}}) \prod_{t=k}^{1} \mathbf{T}_{i_t j_t}^{-1} \Bigg\|_F^2 \\
= & \| \mathbf{C} -  \mathbf{T}_{i_k j_k} \mathbf{B}^{(k)} \mathbf{T}_{i_k j_k}^{-1} \|_F^2 \\
= & \|\mathbf{C} \|_F^2 + \| \mathbf{B}^{(k)} \|_F^2 - 2\text{tr}(\mathbf{C}^T \mathbf{B}^{(k)}) + \mathscr{C}_{i_k j_k}^{(f_k)}(a_k),
\end{aligned}
\end{equation}
with the matrix
\begin{equation}
\mathbf{B}^{(k)} = \left( \prod_{t=1}^{k-1} \mathbf{T}_{i_t j_t} \right) \text{diag}(\mathbf{\bar{c}}) \left( \prod_{t=k-1}^{1} \mathbf{T}_{i_t j_t}^{-1} \right),
\end{equation}
where we have defined the cost values as
\begin{equation}
\begin{aligned}
\mathscr{C}_{i_k j_k}^{(1)}&(a_k) =
(a_k-1)^2 N_{i_k}^{(k)} + (a_k^{-1} - 1)^2M_{i_k}^{(k)} \\
& \ - a_k^{-2}(a_k-1)^2(a_k^2 +1) L_{i_k}^{(k)} - 2(a_k-1)V_{i_k i_k}^{(k)} \\
& \ - 2(a_k^{-1}-1)H_{i_k i_k}^{(k)} + 2a_k^{-1}(a_k-1)^2J_{i_k i_k}^{(k)},
\end{aligned}
\label{eq:c1}
\end{equation}
\begin{equation}
\begin{aligned}
&\mathscr{C}_{i_k j_k}^{(2)}(a_k) =
a_k^2 (N_{j_k}^{(k)} - L_{j_k}^{(k)} + M_{i_k}^{(k)} - L_{i_k}^{(k)}) \\
& \ + a_k^2(B_{j_k j_k}^{(k)} - B_{i_k i_k}^{(k)} - a_kB_{j_k i_k}^{(k)})^2 \\
& \ -2a_kV_{i_k j_k}^{(k)} + 2a_kH_{j_k i_k}^{(k)} + 2a_k^2 B_{j_k i_k}^{(k)}(C_{i_k j_k} - B_{i_k j_k}^{(k)}),
\end{aligned}
\label{eq:c2}
\end{equation}
\begin{equation}
\begin{aligned}
& \mathscr{C}_{i_k j_k}^{(3)}(a_k) =
a_k^2 (N_{i_k}^{(k)} - L_{j_k}^{(k)} + M_{j_k}^{(k)} - L_{i_k}^{(k)}) \\
& \ + a_k^2(B_{i_k i_k}^{(k)} - B_{j_k j_k}^{(k)} - a_kB_{i_k j_k}^{(k)})^2 \\
& \ -2a_k V_{j_k i_k}^{(k)} + 2a_k H_{i_k j_k}^{(k)} + 2a_k^2 B_{i_k j_k}^{(k)}(C_{j_k i_k} - B_{j_k i_k}^{(k)}),
\end{aligned}
\label{eq:c3}
\end{equation}
and we have used the quantities:
\begin{equation}
\begin{aligned}
\mathbf{V}^{(k)} = (\mathbf{C} - \mathbf{B}^{(k)}){\mathbf{B}^{(k)}}^T, & \ \mathbf{H}^{(k)} = (\mathbf{C} - \mathbf{B}^{(k)})^T \mathbf{B}^{(k)}, \\
\mathbf{J}^{(k)} \! = \! (\mathbf{C} - \mathbf{B}^{(k)}) \odot \mathbf{B}^{(k)}, & \ L_{i_k} = (B_{i_k i_k}^{(k)})^2, \\
N_{i_k} = \| \mathbf{B}^{(k)}_{i_k, :} \|_2^2,& \ M_{i_k} = \| \mathbf{B}^{(k)}_{:, i_k} \|_2^2.
\end{aligned}
\end{equation}
The goal is to search for
\begin{equation}
(f_k^\star, i_k^\star, j_k^\star, a_k^\star) = \underset{f_k \in \{1,2,3 \},\ j_k > i_k}{\arg \min} \mathscr{C}_{i_k j_k}^{(f_k)}(a_k).
\end{equation}
Because all three $\mathscr{C}_{i_k j_k}^{(k)}(a_k)$ are polynomials in $a_k$ of degree four and five their minimization therefore reduces to finding the roots of their derivatives and evaluating the values at those points searching for the minimum value.\\
We have used the following explicit inverse formulas:
\begin{equation}
\begin{bmatrix} 1 & 0 \\ a & 1\end{bmatrix}^{-1} \! \! \! \! =  \begin{bmatrix} 1 & 0 \\ -a & 1\end{bmatrix} \text{ and } \begin{bmatrix} 1 & a \\ 0 & 1\end{bmatrix}^{-1} \! \! \! \! =  \begin{bmatrix} 1 & -a \\ 0 & 1\end{bmatrix}.
\end{equation}

\subsection*{Proof of Theorem 4}

The expression \eqref{eq:unsymmetricWithT} can be developed to
\begin{equation}
\begin{aligned}
\| & \mathbf{C} - \mathbf{A}^{(k)} \mathbf{T}_{i_k j_k} \mathbf{B}^{(k)} \mathbf{T}_{i_k j_k}^{-1} \mathbf{D}^{(k)}  \|_F^2 \\
= & \| \text{vec}(\mathbf{C}) - ({\mathbf{D}^{(k)}}^T \mathbf{T}_{i_k j_k}^{-T}  \otimes \mathbf{A}^{(k)} \mathbf{T}_{i_k j_k} ) \text{vec}(\mathbf{B}^{(k)}) \|_F^2 \\
= & \| \text{vec}(\mathbf{C}) - ({\mathbf{D}^{(k)}}^T \! \! \! \! \!  \otimes \! \mathbf{A}^{(k)} \! ) (\mathbf{T}_{i_k j_k}^{-T} \! \! \otimes \! \mathbf{T}_{i_k j_k} \! ) \text{vec}(\mathbf{B}^{(k)}) \|_F^2 \\
= & \| \text{vec}(\mathbf{C}) - [\text{vec}(\mathbf{B}^{(k)})^T \otimes ({\mathbf{D}^{(k)}}^T  \otimes \mathbf{A}^{(k)}  )] \\
&    \quad \quad \quad \quad \quad \quad \quad \quad \quad \quad \quad \quad  \text{vec}(\mathbf{T}_{i_k j_k}^{-T} \otimes \mathbf{T}_{i_k j_k} )  \|_F^2 \\
= & \| \mathbf{w} - \mathbf{Px} \|_F^2 = \mathbf{w}^T \mathbf{w} + \mathscr{D}_{i_k j_k}^{(f_k)}(a_k),
\end{aligned}
\label{eq:developmentforT}
\end{equation}
where we have used twice the fact that  $(\mathbf{A} \otimes \mathbf{C})(\mathbf{B} \otimes \mathbf{D}) = (\mathbf{AB} \otimes \mathbf{CD})$, twice that $\text{vec}(\mathbf{ABC}) = (\mathbf{C}^T \otimes \mathbf{A})\text{vec}(\mathbf{B})$ and $\mathbf{x} \in \mathbb{R}^2$ is a function only of $a_k$. We have defined $\mathbf{w} = \text{vec}(\mathbf{C}) - \sum_i \sum_j B^{(k)}_{j i} (\mathbf{D}^{(k)}_{i,:} \otimes \mathbf{A}^{(k)}_{:,j})$. But this large vector of size $n^2$ and other large matrices (Kronecker products of size $n^2$) are never explicitly constructed but the objective function is minimized by exploiting the structures \eqref{eq:Ttransform} and \eqref{eq:localstructure2}. The cost values for the scaling and the two shears respectively are
\begin{equation}
\begin{aligned}
\mathscr{D}_{i_k j_k}^{(1)}(a_k) = & (a_k - 1)^2R_{11}^{(1)} + a_k^{-2}(1-a_k)^2R_{22}^{(1)} \\
& - 2a_k^{-1}(a_k-1)^2R_{12}^{(1)} \\
& - 2(a_k-1)g_1^{(1)} - 2a_k^{-1}(1-a_k)g_2^{(1)},
\end{aligned}
\label{eq:newc1}
\end{equation}
\begin{equation}
\begin{aligned}
\mathscr{D}_{i_k j_k}^{(f_k)}(a_k) = & a_k^4 R_{22}^{(f_k)} + 2a_k^3R_{12}^{(f_k)} + a_k^2(R_{11}^{(f_k)} \\
&  + 2g_2^{(f_k)}) + 2a_kg_1^{(f_k)},\ f_k \in \{2, 3\}.
\end{aligned}
\label{eq:newc2andc3}
\end{equation}
For completeness, we give the explicit formulas for all the quantities in \eqref{eq:newc1} and \eqref{eq:newc2andc3}
\begin{itemize}
    \item for $f_k = 1$ we have:\newline
    $R_{11}^{(1)} = H_{i_k i_k}^{(k)}\sum_{i \neq i_k}B_{i_k i}^{(k)}  \sum_{j \neq i_k} B_{i_k j}^{(k)} V_{ji}^{(k)} $, $R_{22}^{(1)} = V_{i_k i_k}^{(k)} \sum_{i \neq i_k} B_{i i_k}^{(k)}  \sum_{j \neq i_k} B_{j i_k}^{(k)} H_{ij}^{(k)} $, $R_{12}^{(1)} = \sum_{i \neq i_k} B_{i_k i}^{(k)} V_{i i_k}^{(k)}   \sum_{j \neq i_k} B_{j i_k}^{(k)} H_{i_k j}^{(k)} $, \newline $g_1^{(1)} \! = \! \! \! \sum_{t \neq i_k} B_{i_k t}^{(k)}  \sum_i \left(D_{ti}^{(k)} J_{i_k i}^{(k)} \! - \! V_{ti}^{(k)}  \sum_{j}  B_{ji}^{(k)} H_{i_k j}^{(k)} \right)$,\newline $g_2^{(1)} \! = \! \! \! \sum_{t \neq i_k} B_{t i_k}^{(k)} \sum_{i} \left(D_{i_k i}^{(k)} J_{t i}^{(k)}  \! - \! V_{i_k i}^{(k)} \sum_j  B_{ji}^{(k)}  H_{j t}^{(k)} \right)$;
    
    \item for $f_k = 2$ we have:\newline
    $R_{11}^{(2)} = \sum_i \sum_j B_{j_k i}^{(k)} B_{j_k j}^{(k)} V_{ij}^{(k)} H^{(k)}_{j_k-i_k+1,j_k-i_k+1} +$ \newline $B_{i i_k}^{(k)} B_{j i_k}^{(k)} V_{j_k j_k}^{(k)} H_{i j}^{(k)} - 2 B_{j_k i}^{(k)} B_{j i_k}^{(k)} V_{j_k i}^{(k)} H^{(k)}_{j_k-i_k+1,j}$,\newline $R_{22}^{(2)} = (B_{j_k i_k}^{(k)})^2 V_{j_k j_k}^{(k)} H^{(k)}_{j_k-i_k+1,j_k-i_k+1}$, \newline $R_{12}^{(2)} = B_{j_k i_k}^{(k)} \left( \sum_i B_{i i_k}^{(k)} V_{j_k j_k}^{(k)} H^{(k)}_{i, j_k-i_k+1} \right. - \left. B^{(k)}_{j_k i} V^{(k)}_{i j_k} H_{j_k-i_k+1,j_k-i_k+1}\right)$,\newline
    $g_{1}^{(2)} = \sum_t \sum_i \left( \sum_j B_{ji}^{(k)}  \left( B_{j_k t}^{(k)} V_{ti}^{(k)} H_{j_k-i_k+1, j} \right. \right. $ \newline \quad \ \ $ \left. \left. - B_{t i_k}^{(k)} V_{j_k i}^{(k)} H_{tj} \! \right) \! + \! B_{t i_k}^{(k)} D_{j_k i}^{(k)} J_{ti}^{(k)} \! \! - \! B_{j_k t}^{(k)} D_{t i}^{(k)} J^{(k)}_{j_k-i_k+1, i} \! \right),$ \newline $g_2^{(2)}\! \! = \! \! B_{j_k i_k}^{(k)} \! \sum_i \! \! \left( \! D^{(k)}_{j_k i}J^{(k)}_{j_k-i_k+1, i} \! \! - \!  \! \! \sum_j \!\! B_{ji}^{(k)} V^{(k)}_{j_k i}  H^{(k)}_{j_k-i_k+1,j} \! \! \right)$;
    
    \item for $f_k = 3$ we have:\newline
    $R_{11}^{(3)} = \sum_i \sum_j B_{i_k i}^{(k)} B_{i_k j}^{(k)} V_{ij}^{(k)} H^{(k)}_{j_k j_k} + B_{i j_k}^{(k)} B_{j j_k}^{(k)} V_{i_k i_k}^{(k)} H_{i j}^{(k)} - 2 B_{i_k i}^{(k)} B_{j j_k}^{(k)} V_{i_k i}^{(k)} H^{(k)}_{j_k j}$, $R_{22}^{(2)} = (B_{i_k j_k}^{(k)})^2 V_{i_k i_k}^{(k)} H^{(k)}_{j_k j_k}$, $R_{12}^{(2)} = B_{i_k j_k}^{(k)} \left( \sum_i B_{i j_k}^{(k)} V_{i_k i_k}^{(k)} H^{(k)}_{i j_k} \right. - \left. B^{(k)}_{i_k i} V^{(k)}_{i i_k} H_{j_k j_k}\right)$,\newline
    $g_{1}^{(2)} = \sum_t \sum_i \left( \sum_j B_{ji}^{(k)}  \left( B_{i_k t}^{(k)} V_{ti}^{(k)} H_{j_k j} \right. \right. $ \newline \quad \ \ $ \left. \left. - B_{t j_k}^{(k)} V_{i_k i}^{(k)} H_{tj} \right)  + B_{t j_k}^{(k)} D_{i_k i}^{(k)} J_{ti}^{(k)} - B_{i_k t}^{(k)} D_{t i}^{(k)} J^{(k)}_{j_k i} \right),$ \newline $g_2^{(2)} = B_{i_k j_k}^{(k)} \sum_i \left(D^{(k)}_{i_k i}J^{(k)}_{j_k i}  -  \sum_j B_{ji}^{(k)} V^{(k)}_{i_k i}  H^{(k)}_{j_k j} \right)$,
\end{itemize}
and we have used the quantities:
\begin{equation}
\mathbf{V}^{(k)} \! = \! \mathbf{D}^{(k)} {\mathbf{D}^{(k)}}^T\!, \mathbf{H}^{(k)} \! = \! {\mathbf{A}^{(k)}}^T \mathbf{A}^{(k)}, \mathbf{J}^{(k)} \! = \! {\mathbf{A}^{(k)}}^T \mathbf{C}.
\end{equation}
Similarly to the initialization step, the goal is to search for
\begin{equation}
(f_k^\star, i_k^\star, j_k^\star, a_k^\star) = \underset{f_k \in \{1,2,3 \},\ j_k > i_k}{\arg \min} \mathscr{D}_{i_k j_k}^{(f_k)}(a_k).
\label{eq:solutionsT}
\end{equation}
Again, all three $\mathscr{D}_{i_k j_k}^{(k)}(a_k)$ are polynomials in $a_k$ of degree four and five their minimization therefore reduces to finding the roots of their derivatives and evaluating the values at those points searching for the minimum value.

\begin{figure*}[!t]
    \centering
    \includegraphics[trim = 10 2 95 10, clip, width=\textwidth]{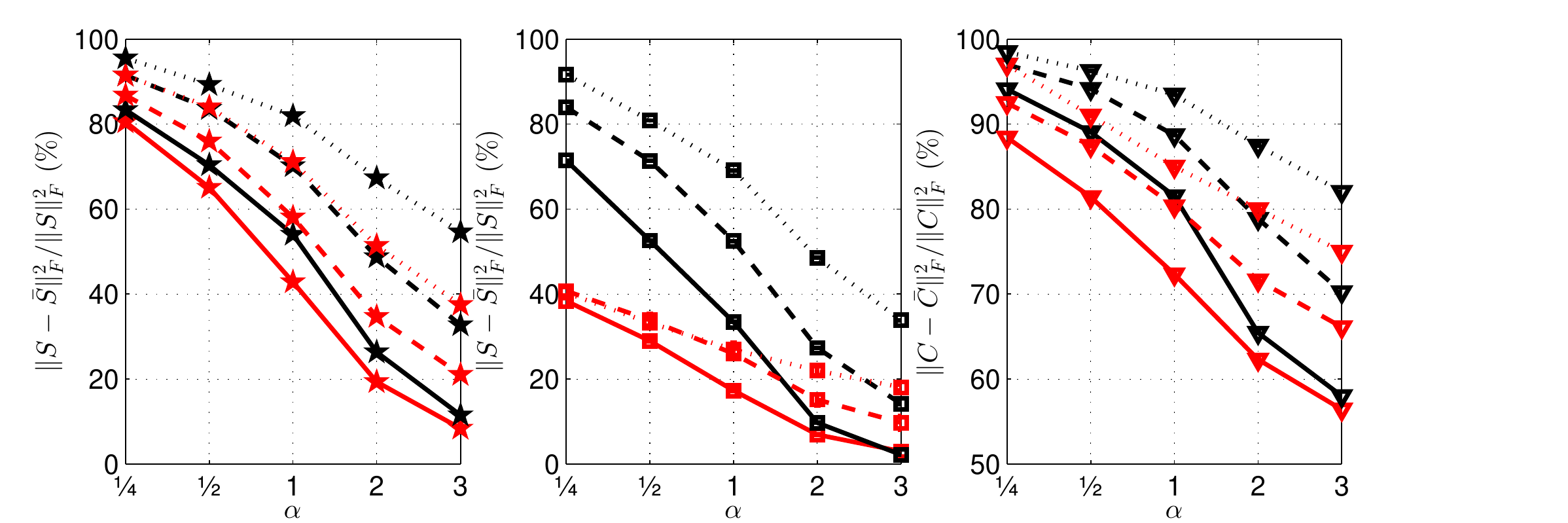}
    \caption{In red, approximation accuracy (mean and std) of the proposed method for randomly generated matrices as a function of the number of transformations $g$ or $m$ going as $\alpha n \log_2 n $. Given a matrix $\mathbf{X}$ with entries i.i.d. standard Gaussian we have results for: symmetric indefinite $\mathbf{S} = \mathbf{X} + \mathbf{X}^T$ (left), symmetric positive semidefinite $\mathbf{S} = \mathbf{XX}^T$ (central) and unsymmetric $\mathbf{C} = \mathbf{X}$ (right). Results are shown for $n = 512$ (dotted), $n=256$ (dashed) and $n=128$ (solid) and all methods update also the spectrum of the estimation. Note that the achieved accuracy is better for the positive definite case. In black, for comparison, the results of $r$-rank approximations: for the symmetric case $r = 3 \alpha n \log_2 n$ while for the unsymmetric case $r=\alpha n \log_2 n$ (for these values we match the numerical complexity of the transformations $\mathbf{\bar{U}}$ and $\mathbf{\bar{T}}$, we count $2rn$ operations for matrix-vector multiplication with the $r$-rank matrix). Results are averaged over 100 realizations.}
    \label{fig:figure1}
\end{figure*}
\begin{figure*}[!t]
    \centering
    \includegraphics[trim = 103 0 100 0, clip, width=\textwidth]{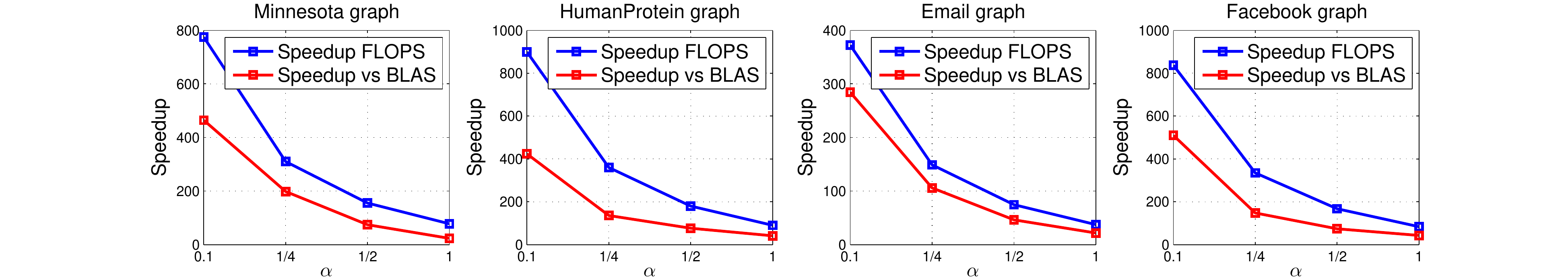}
    \caption{Average speedup achieved for matrix-vector multiplication between the full eigenspaces versus their approximations using Algorithm 1 for the graphs from Figure \ref{fig:figure3}. We show the FLOP count (count of the number of operations: $6\alpha n \log_2 n$ for the G-transformations and $2 \alpha n \log_2 n$ for the T-transformations) as compared to regular matrix-vector multiplication ($2n^2$) and the actual matrix-vector multiplication runtime as compated to the LAPACK, Level 2 BLAS, implementation (SGEMV). Application of the butterflies is implemented in the C programming language (scripting languages such as Matlab perform very poorly if all $g$ or $m$ basic transformations are applied sequentially). No parallelism is used in any of these experiments. Code runs on a 2.3GHz Quad-Core Intel Core i5 system with 16GB LPDDR3 memory.}
    \label{fig:figure6}
\end{figure*}    
\subsection*{Additional experimental results}

In Figure \ref{fig:figure1}, we show the accuracy of the approximation for randomly generated symmetric (both positive definite and indefinite), and general matrices. Details about the experimental setup are given in the figure caption. The proposed algorithm uses again only a polishing step and not a full transform update, for computational efficiency.

Finally, in Figure \ref{fig:figure6} we show the speedup achieved by the proposed transformations. Throughout the paper, we define numerically efficient transformations as those that present a low number of additions and multiplications in matrix-vector operations, i.e., FLOP (floating-point operations) count. In this figure, we also show the speedup in terms of actual running time and compare it with the FLOP count. We stress that the speedup does not refer to the running time of the proposed Algorithm 1, but the application of the transformations reached by this algorithm. For this figure, the butterfly transformations (G and T transformations) are implemented in the C programming language (a Matlab implementation of the application of these butterflies is hopelessly slow as compared to just matrix-vector multiplication in Matlab, i.e., the ``*'' operation calls compiled BLAS functions as opposed to parsing and running Matlab scripts).

\newpage
\bibliography{refs}

\begin{thebibliography}{27}
\providecommand{\natexlab}[1]{#1}
\providecommand{\url}[1]{\texttt{#1}}
\expandafter\ifx\csname urlstyle\endcsname\relax
  \providecommand{\doi}[1]{doi: #1}\else
  \providecommand{\doi}{doi: \begingroup \urlstyle{rm}\Url}\fi

\bibitem[Cooley and Tukey(1965)]{FFT}
J.~W. Cooley and J.~W. Tukey.
\newblock An algorithm for the machine calculation of complex {Fourier} series.
\newblock \emph{Math. Comp.}, 19:\penalty0 297--301, 1965.

\bibitem[Daubechies and Sweldens(1998)]{Lifting2}
I.~Daubechies and W.~Sweldens.
\newblock Factoring wavelet transforms into lifting steps.
\newblock \emph{J. Fourier Anal. App.}, 4\penalty0 (3):\penalty0 247--269,
  1998.

\bibitem[Defferrard et~al.(2015)Defferrard, Martin, Pena, and Perraudin]{pygsp}
M.~Defferrard, L.~Martin, R.~Pena, and N.~Perraudin.
\newblock {PyGSP}: {G}raph {S}ignal {P}rocessing in {P}ython, 2015.

\bibitem[Frerix and Bruna(2019)]{Frerix2019ApproximatingOM}
T.~Frerix and J.~J. Bruna.
\newblock Approximating orthogonal matrices with effective {Givens}
  factorization.
\newblock In \emph{Proceedings 36th International Conference on Machine
  Learning (ICML)}, 2019.

\bibitem[Gander et~al.(1989)Gander, Golub, and von Matt]{SphereLS}
W.~Gander, G.~H. Golub, and U.~von Matt.
\newblock A constrained eigenvalue problem.
\newblock \emph{Linear Algebra Appl.}, 114-115:\penalty0 815--839, 1989.

\bibitem[Givens(1958)]{GivensRotations}
W.~Givens.
\newblock Computation of plain unitary rotations transforming a general matrix
  to triangular form.
\newblock \emph{Journal of the Society for Industrial and Applied Mathematics},
  6\penalty0 (1):\penalty0 26--50, 1958.

\bibitem[Golub and van~der Vorst(2000)]{DecompositionsEig}
G.~H. Golub and H.~A. van~der Vorst.
\newblock Eigenvalue computation in the 20th century.
\newblock \emph{Journal of Computational and Applied Mathematics}, 123\penalty0
  (1-2):\penalty0 35--65, 2000.

\bibitem[Golub and van Loan(1996)]{Golub1996}
G.~H. Golub and C.~F. van Loan.
\newblock \emph{Matrix Computations}.
\newblock Johns Hopkins University Press, 1996.

\bibitem[Guimerà et~al.(2004)Guimerà, Danon, Diaz-Guilera, Giralt, and
  Arenas]{Guimera2004}
R.~Guimerà, L.~Danon, A.~Diaz-Guilera, F.~Giralt, and A.~Arenas.
\newblock Self-similar community structure in a network of human interactions.
\newblock \emph{Physical review. E, Statistical, nonlinear, and soft matter
  physics}, 68:\penalty0 065103, 2004.

\bibitem[Hardy et~al.(1952)Hardy, Littlewood, and Polya]{Hardy}
G.~H. Hardy, J.~E. Littlewood, and G.~Polya.
\newblock Cambridge University Press, 1952.

\bibitem[Henrici(1958)]{10.2307/2098731}
P.~Henrici.
\newblock On the speed of convergence of cyclic and quasicyclic {Jacobi}
  methods for computing eigenvalues of {Hermitian} matrices.
\newblock \emph{Journal of the Society for Industrial and Applied Mathematics},
  6\penalty0 (2):\penalty0 144--162, 1958.

\bibitem[Jacobi(1846)]{JacobiProcess}
C.~Jacobi.
\newblock Uber ein leichtes {Verfahren} die in der {Theorie} der
  {Sacularstorungen} vorkommenden {Gleichungen} numerisch aufzulosen.
\newblock \emph{Journal fur die reine und angewandte {Mathematik}},
  30:\penalty0 51--94, 1846.

\bibitem[Kondor et~al.(2014)Kondor, Teneva, and Garg]{Kondor2014MMF}
R.~Kondor, N.~Teneva, and V.~K. Garg.
\newblock Multiresolution matrix factorization.
\newblock In \emph{Proceedings 31st International Conference on Machine
  Learning (ICML)}, pages II--1620--II--1628, 2014.

\bibitem[{Kyng} and {Sachdeva}(2016)]{AGE}
R.~{Kyng} and S.~{Sachdeva}.
\newblock Approximate {Gaussian} elimination for {Laplacians} - fast, sparse,
  and simple.
\newblock In \emph{IEEE 57th Annual Symposium on Foundations of Computer
  Science (FOCS)}, pages 573--582, 2016.

\bibitem[Le~Magoarou et~al.(2018)Le~Magoarou, Gribonval, and
  Tremblay]{lemagoarou:hal-01416110}
L.~Le~Magoarou, R.~Gribonval, and N.~Tremblay.
\newblock {Approximate fast graph Fourier transforms via multi-layer sparse
  approximations}.
\newblock \emph{IEEE Transactions on Signal and Information Processing over
  Networks}, 4\penalty0 (2):\penalty0 407--420, 2018.

\bibitem[Lee et~al.(2008)Lee, Nadler, and Wasserman]{Treelets}
A.~B. Lee, B.~Nadler, and L.~Wasserman.
\newblock Treelets - an adaptive multi-scale basis for sparse unordered data.
\newblock \emph{Annals of Applied Statistics}, 2\penalty0 (2):\penalty0
  435--471, 2008.

\bibitem[Leskovec and Mcauley(2012)]{NIPS2012_4532}
J.~Leskovec and J.~J. Mcauley.
\newblock Learning to discover social circles in ego networks.
\newblock In \emph{Advances in Neural Information Processing Systems 25}, pages
  539--547. 2012.

\bibitem[Meijerink and Vorst(1977)]{ILU}
J.~A. Meijerink and H.~A. Vorst.
\newblock An iterative solution method for linear systems of which the
  coefficient matrix is a symmetric {M}-matrix.
\newblock \emph{Math. Comp.}, 31:\penalty0 148--162, 1977.

\bibitem[Mudrakarta et~al.(2019)Mudrakarta, Trivedi, and
  Kondor]{mudrakarta2019asymmetric}
P.~K. Mudrakarta, S.~Trivedi, and R.~Kondor.
\newblock Asymmetric multiresolution matrix factorization.
\newblock \emph{arXiv 1910.05132}, 2019.

\bibitem[Rual et~al.(2005)Rual, Venkatesan, Hao, Hirozane-Kishikawa, Dricot,
  Li, Berriz, Gibbons, Dreze, Ayivi-Guedehoussou, Klitgord, Simon, Boxem,
  Milstein, Rosenberg, Goldberg, Zhang, Wong, Franklin, and Vidal]{Rual2005}
J.-F. Rual, K.~Venkatesan, T.~Hao, T.~Hirozane-Kishikawa, A.~Dricot, N.~Li,
  G.~Berriz, F.~Gibbons, M.~Dreze, N.~Ayivi-Guedehoussou, N.~Klitgord,
  C.~Simon, M.~Boxem, S.~Milstein, J.~Rosenberg, D.~Goldberg, L.~Zhang,
  S.~Wong, G.~Franklin, and M.~Vidal.
\newblock Towards a proteome-scale map of the human protein-protein interaction
  network.
\newblock \emph{Nature}, 437:\penalty0 1173--8, 2005.

\bibitem[Rusu(2018)]{rusu2018learning}
C.~Rusu.
\newblock Learning multiplication-free linear transformations.
\newblock \emph{arXiv 1812.03412}, 2018.

\bibitem[Rusu and Rosasco(2019)]{FastPCA}
C.~Rusu and L.~Rosasco.
\newblock Fast approximation of orthogonal matrices and application to {PCA}.
\newblock \emph{arXiv 1907.08697}, 2019.

\bibitem[Rusu and Thompson(2017)]{FastSparsifyingTransforms}
C.~Rusu and J.~Thompson.
\newblock Learning fast sparsifying transforms.
\newblock \emph{IEEE Trans. Sig. Proc.}, 65\penalty0 (16):\penalty0 4367--4378,
  2017.

\bibitem[Schonemann(1968)]{Schonemann1968}
P.~Schonemann.
\newblock On two-sided orthogonal {Procrustes} problems.
\newblock \emph{Psychometrika}, 33\penalty0 (1):\penalty0 19--33, 1968.

\bibitem[Shabat et~al.(2018)Shabat, Shmueli, Aizenbud, and Averbuch]{RLU}
G.~Shabat, Y.~Shmueli, Y.~Aizenbud, and A.~Averbuch.
\newblock Randomized {LU} decomposition.
\newblock \emph{Applied and Computational Harmonic Analysis}, 44\penalty0
  (2):\penalty0 246 -- 272, 2018.

\bibitem[Shalit and Chechik(2014)]{Shalit}
U.~Shalit and G.~Chechik.
\newblock Coordinate-descent for learning orthogonal matrices through {Givens}
  rotations.
\newblock In \emph{Proceedings 31st International Conference on Machine
  Learning (ICML)}, pages I--548--I--556, 2014.

\bibitem[{Stewart}(2000)]{814658}
G.~W. {Stewart}.
\newblock The decompositional approach to matrix computation.
\newblock \emph{Computing in Science Engineering}, 2\penalty0 (1):\penalty0
  50--59, 2000.

\end{thebibliography}
\bibliographystyle{abbrvnat}

\newpage

\end{document}